\documentclass[review,3p,onecolumn,times,article]{elsarticle}

\usepackage{amssymb}
\usepackage{amsmath}
\usepackage{lineno}
\usepackage{longtable}
\usepackage{arydshln}
\usepackage{booktabs}
\usepackage{multirow}

\usepackage{bm}
\usepackage{setspace}
\usepackage{makecell}

\usepackage{orcidlink}

\usepackage{hyperref}
\hypersetup{colorlinks=true, citecolor=blue, linkcolor=blue, urlcolor=blue}

\date{}
\journal{}
\makeatletter
\def\ps@pprintTitle{%
 \let\@oddhead\@empty
 \let\@evenhead\@empty
 \let\@oddfoot\@empty
 \let\@evenfoot\@empty
}
\makeatother

\begin{document}

\begin{frontmatter}



\title{Advancements in Chinese font generation since deep learning era: A survey}



\author[1]{Weiran Chen}
\ead{wrchen2023@stu.suda.edu.cn}


\author[1]{Guiqian Zhu}


\author[1]{Ying Li}


\author[1]{Yi Ji}


\author[1]{Chunping Liu\corref{cor1}}
\ead{cpliu@suda.edu.cn}

\cortext[cor1]{Corresponding author}

\address[1]{School of Computer Science and Technology, Soochow University, 215006, Suzhou, China}

\begin{abstract}
Chinese font generation aims to create a new Chinese font library based on some reference samples. It is a topic of great concern to many font designers and typographers. Over the past years, with the rapid development of deep learning algorithms, various new techniques have achieved flourishing and thriving progress. Nevertheless, how to improve the overall quality of generated Chinese character images remains a tough issue. In this paper, we conduct a holistic survey of the recent Chinese font generation approaches based on deep learning. To be specific, we first illustrate the research background of the task. Then, we outline our literature selection and analysis methodology, and review a series of related fundamentals, including classical deep learning architectures, font representation formats, public datasets, and frequently-used evaluation metrics. After that, relying on the number of reference samples required to generate a new font, we categorize the existing methods into two major groups: many-shot font generation and few-shot font generation methods. Within each category, representative approaches are summarized, and their strengths and limitations are also discussed in detail. Finally, we conclude our paper with the challenges and future directions, with the expectation to provide some valuable illuminations for the researchers in this field.
\end{abstract}




\end{frontmatter}




\section{Introduction}\label{Introduction}
Chinese characters are the direct inheritors of 5,000 years Chinese civilization. They have both profound cultural connotations and artistic value~\cite{1}. As a vital medium for information transmission, Chinese characters play an indispensable role in people's work and lives. Nowadays, with the booming progress of Internet technology, the demands for diverse and customized Chinese fonts grow rapidly across various applications, such as logo design, ancient character restoration, handwriting imitation, digital media creation and so on. Thus, Chinese font generation has become a crucial research topic. However, different from other languages, Chinese characters are not only vast in number but also possess highly intricate structures. The diversity and structural complexity of characters make Chinese font generation a particularly challenging task because each Chinese character's unique form needs to be precisely captured. Consequently, Chinese font design is a labor-intensive and time-consuming process, which requires meticulous handcrafting by proficient experts.

During the past years, to alleviate the dependence on manual efforts, more and more specialists have focused on automatic Chinese font generation. Automatic Chinese font generation is essentially an imitation task, which refers to utilizing visual computing techniques to generate a new font that mimics the style of reference images while preserving the content from source images~\cite{2}. In the early stage, people mainly utilize traditional methods to realize Chinese font generation. These methods can be broadly divided into two categories, which are component-based methods and morphology-based methods respectively. Both of them, component-based methods often separate Chinese characters into different radicals or strokes, and then recombine them to generate new characters~\cite{3,4}. Conversely, morphology-based methods typically create new fonts through modelling and analyzing the shape and line structure of Chinese characters, such as the skeleton or contour~\cite{5,6}. While these traditional methods have yielded certain effects, their generated results are notably influenced by prior knowledge. In addition, due to the reliance on fixed rules and patterns, these methods often lead to a lack of stylistic diversity in the generated fonts~\cite{7}. 

Later, in order to overcome the deficiencies of traditional methods, researchers are increasingly turning to adopt deep learning techniques for the Chinese font generation task. Compared with traditional models, deep learning excels in discovering the inherent laws of sample data and synthesizing low-level features into more high-level feature representations~\cite{8,9}. It has great advantages in both feature extraction and model fitting. As of now, in numerous research publications, many scholars have leveraged the powerful deep learning models in innovative ways~\cite{10, 11, 12, 13} to improve the final results' quality. Nevertheless, the broad array of methods makes it difficult for academics to choose the most appropriate one. Therefore, there is substantial value in systematically reviewing these methods and organizing them into distinct classes.

Currently, there have been several survey papers on Chinese font generation task~\cite{7,14,15,16,17,170,171}. Whereas, some of them focus narrowly on specific aspects, such as GAN-based methods~\cite{14} or many-shot generation techniques~\cite{17}, without paying attention to the full spectrum of approaches. Besides, several surveys~\cite{15,16} attempt to cover a wider range, but they still fall short in capturing the diversity of available methods. Along with that, a recent survey categorizes the generation algorithms into stroke-trajectory-based algorithms and glyph-image-based algorithms according to the type of input data~\cite{7}. While this classification is simple and straightforward, it does not adequately represent the variety of Chinese font generation methods. In addition, some newly-published surveys~\cite{170,171} either lack in-depth comparisons between different methods or narrowly concentrate on bitmap font generation. Furthermore, these previous reviews are limited over the time period before July 2024, thereby missing out on the cutting-edge advancements in this swiftly evolving domain.

\begin{figure}[h]%
\centering
\includegraphics[width=0.85\textwidth]{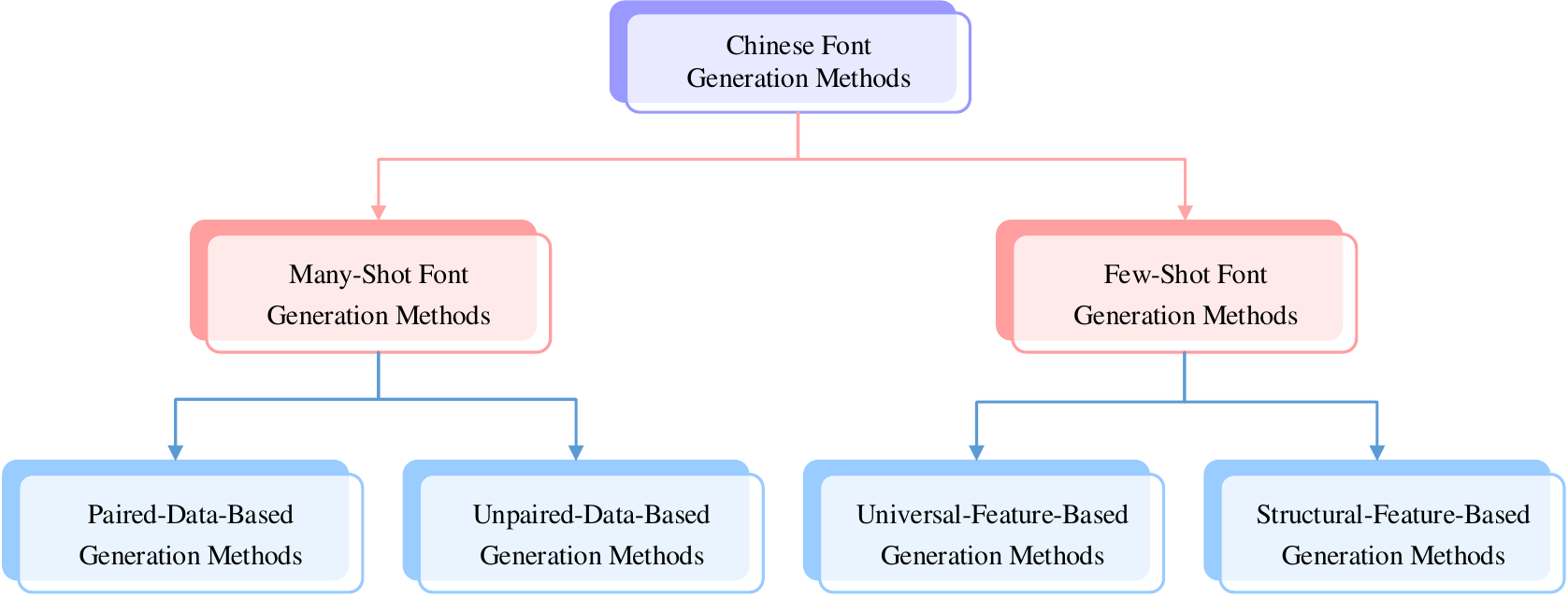}
\caption{Taxonomy of Chinese font generation methods based on deep learning.}\label{Taxonomy}
\end{figure}

\begin{figure}[h]%
\centering
\includegraphics[width=\textwidth]{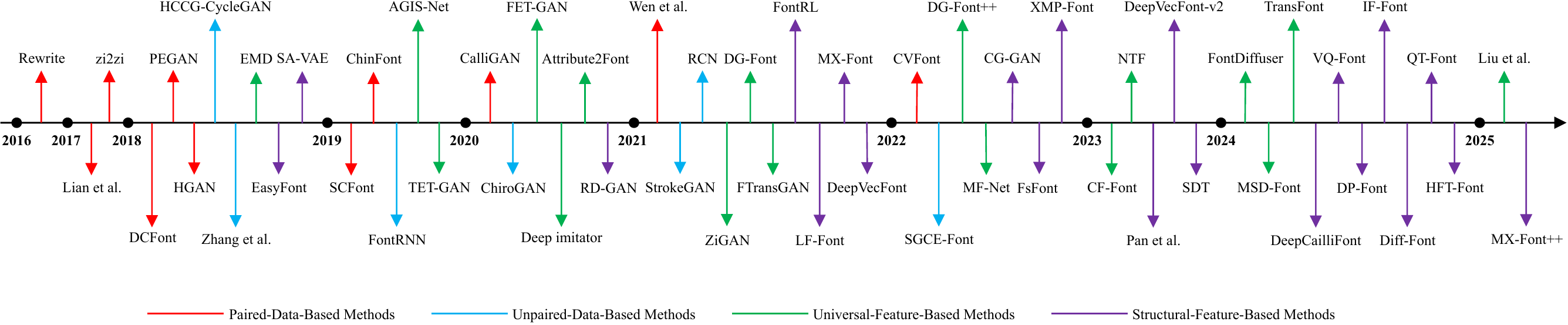}
\caption{Milestones and significant works of deep-learning-based Chinese font generation methods.}\label{Milestones}
\end{figure}

In this paper, we present a comprehensive and up-to-date survey on the Chinese font generation methods since the deep learning era. Building on our findings, we propose a novel taxonomy of these generation methods, as shown in Figure~\ref{Taxonomy}. To be specific, here we divide the generation methods into two groups: many-shot font generation methods and few-shot font generation methods. Both them, many-shot font generation methods are further categorized into paired-data-based and unpaired-data-based methods, while few-shot font generation methods are deeply classified into universal-feature-based and structural-feature-based methods. Based on this context, a considerable number of papers have been reviewed to analyze the overall characteristics of those generation algorithms. Some milestones and significant works are presented in Figure~\ref{Milestones}. Our survey provides a thorough examination of various generation methods, which contributes to a better understanding of the current research trend. 

The remainder of this survey is organized as follows: In Section~\ref{Literature collection and coverage}, we describe our literature selection and analysis process briefly. In Section~\ref{Fundamentals}, we introduce some fundamentals of the Chinese font generation task, encompassing several deep learning architectures, font representation formats, public datasets, and evaluation metrics. In Section~\ref{Many-shot Chinese font generation approaches}, we illustrate the many-shot Chinese font generation approaches. In Section~\ref{Few-shot Chinese font generation approaches}, we present the few-shot Chinese font generation approaches. In Section~\ref{Discussion}, we discuss the challenges that are still remaining and propose future research directions. Finally, in Section~\ref{Conclusion}, we draw our conclusions.

\section{Literature collection and coverage}\label{Literature collection and coverage}

It is well recognized that literature collection, retrieval, and screening are the pivotal steps in conducting a systematic review~\cite{18}. Therefore, our research methodology employs a mixed-method review strategy, aligning with established literature in this field to synthesize related findings comprehensively. In particular, we meticulously select phrases like "font generation", "Chinese character style transfer”, "font synthesis", "Chinese calligraphy generation", "character rending", "font style translation" and "character glyph synthesis" as our search keywords. Meanwhile, we utilize a diverse range of reputable academic databases and literature resources for retrieving pertinent papers, which include but are not limited to Science Direct, IEEE Xplore, Springer Link, Web of Science, ACM Digital Library, Google Scholar, CNKI and other authoritative platforms. Besides, we also manually search for relevant work and incorporate some additional papers into our investigation. To be noticed, the time for our literature clustering is adjusted to nearly 10 years, covering the studies published from 2016 to 2025.

Then, via scrutinizing important details like the title and abstract of each publication, we carefully screen all the collected papers and adopt two main exclusion criteria to facilitate our review process. Firstly, we assess the relevance of the research question. Given our focus on Chinese font generation, we exclude the publications that do not involve this area, such as those centred on Latin font, digital font and other language generation tasks. Secondly, we eliminate papers that do not delve deeply into the research topic. This contains those that provide only a superficial analysis and lack in-depth explorations of algorithms as well as experimental results. Through these steps, we strive to uphold rigour and transparency, thereby furnishing a solid research foundation and robust data support for our review.

\begin{figure}[h]%
\centering
\includegraphics[width=0.48\textwidth]{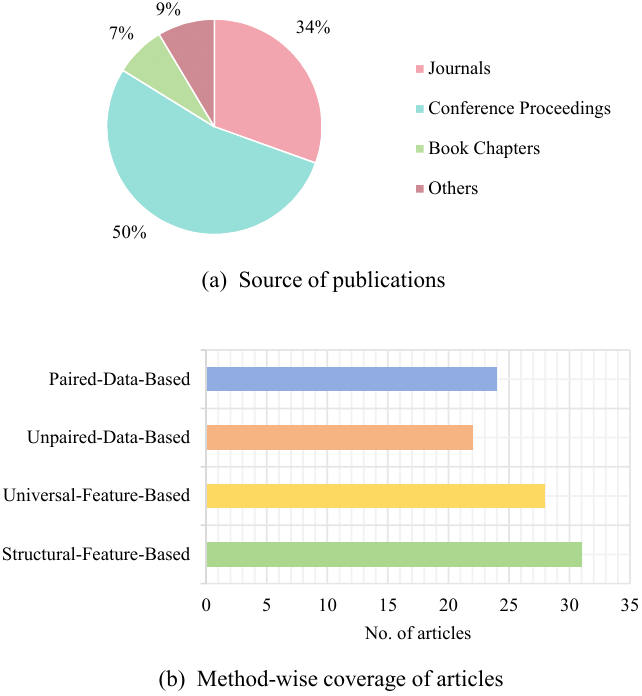}
\caption{Overall distribution of literature in the present survey.}\label{distribution}
\end{figure}

The distribution of publication sources is illustrated in Figure~\ref{distribution}(a), encompassing journals, conference proceedings, book chapters and other transactions from diverse repositories. Besides, as shown in Figure~\ref{distribution}(b), the number of studies is relatively balanced across the four representative types of methods. To be noticed, few-shot font generation methods are slightly more prevalent than many-shot methods. This indicates a growing research interest in data-efficient generation paradigms.

\section{Fundamentals}\label{Fundamentals}
\subsection{Classical deep learning architectures}
In this section, we briefly introduce five foundational architectures that play a crucial role in this domain: Convolutional Neural Network (CNN), Auto-Encoder (AE), Generative Adversarial Network (GAN), Transformer and Diffusion model. These architectures offer varying approaches and capabilities, which conduce to the progress of Chinese font generation methodologies. 

\subsubsection{Convolutional neural network}
CNN is a specialized type of deep neural network, which is often composed of convolutional layers interleaved with pooling layers~\cite{19}. It is particularly designed to analyze grid-like data structures such as images or videos. CNN can efficiently extract spatial hierarchies of data representations through the convolutional and pooling layers. It possesses several predominant characteristics, such as local connections, shared weights, and so on. These properties enable CNN to encapsulate image features while minimizing the information loss during dimensional reconstruction~\cite{15}. In the task of Chinese font generation, CNN is widely used in image preprocessing and feature extraction. However, relying solely on CNN makes it difficult to capture the global structure and intricate stroke details fully.

\subsubsection{Auto-encoder}
As a typical deep unsupervised learning model, AE can automatically learn efficient and compact feature representations from unlabeled data. It intends to realize abstract feature learning by setting the network's desired output equal to the input~\cite{22}. The basic AE architecture usually consists of two essential components: the encoder, which compresses the input data into a concise encoded form, and the decoder, which attempts to reconstruct the original data from the reduced form~\cite{20,21}. By training to narrow the difference between the input and the output, AE can effectively capture the vital features of the data while minimizing noise and redundancy. At present, several variants of AE, like Variational Auto-Encoder (VAE)~\cite{23}, have already been applied to Chinese font generation~\cite{11,24}. Nevertheless, they sometimes suffer from image blurriness and artifacts.
 
\subsubsection{Generative adversarial network}
GAN is a framework for estimating generative models through adversarial training, first introduced by Goodfellow et al.~\cite{25}. It generally comprises two neural networks, the generator and the discriminator, that engage in a zero-sum game to reach an optimal solution. Here, the generator's goal is to produce fake images as realistically as possible based on the given input noise to deceive the discriminator, while the discriminator's role is to distinguish between the real images and generated images. Different from other generative models, GAN does not rely on any prior assumption. It can gradually approximate any probability distribution and generate images without a large number of labelled samples. However, the flexibility in GAN's generation process makes it susceptible to some issues like mode collapse, gradient vanishing and training instability. To alleviate these problems, several variants of GAN, including Conditional Generative Adversarial Network (CGAN)~\cite{26}, Cycle-consistent Generative Adversarial Network (CycleGAN)~\cite{27}, Wasserstein Generative Adversarial Network (WGAN)~\cite{28}, Wasserstein GAN with Gradient Penalty (WGAN-GP)~\cite{49} and StarGAN~\cite{93}, have been proposed to cement the model stability and the quality of generated results. These novel architectures are also extensively employed in the Chinese font generation task~\cite{12,30,31,97}, which demonstrate their efficacy in coping with complex Chinese characters.

\subsubsection{Transformer}
Transformer~\cite{118}, originally proposed for sequence processing, excels at capturing long-range dependencies through the attention mechanism and efficient parallelization. With the emergence of Vision Transformer (ViT)~\cite{138}, which divides images into patches and processes them via standard Transformer encoders, it has demonstrated strong performance on various vision studies. Due to the unified design for both sequential modelling and visual feature extraction, Transformer has increasingly been adopted as a backbone in the Chinese font generation task~\cite{132,161}.

\subsubsection{Diffusion model}
Diffusion model is a neoteric powerful generative approach that leverages an iterative reverse diffusion process to generate high-quality images, which is initially proposed in Sohl-Dickstein et al.~\cite{32} and subsequently improved by Denoising Diffusion Probabilistic Model (DDPM)~\cite{33}. It works by gradually reversing a predefined noise process and transforming the noise distribution into the target data distribution via a series of small and incremental steps. At each step, the model learns to predict and remove the noise, thereby improving the quality and fidelity of the output inch by inch. This constant iterative refinement allows the diffusion model to capture the elaborate details and produce superior outputs. Nonetheless, the application of diffusion models to Chinese font generation remains in its infancy, with considerable potential yet to be explored. Moreover, their inference tends to be slow and computationally intensive, particularly for high-resolution and structurally complex glyphs.

\begin{figure}[h]%
\centering
\includegraphics[width=0.4\textwidth]{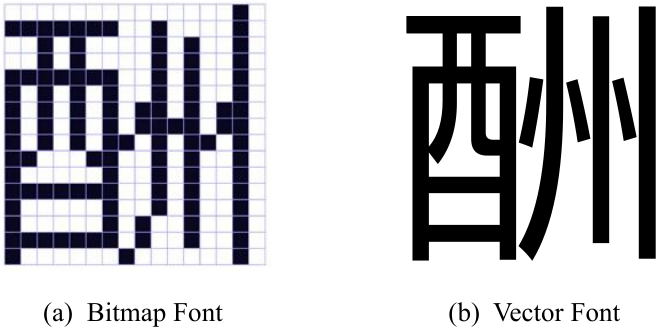}
\caption{Two different font representation formats.}\label{Font Representation}
\end{figure}

\subsection{Font representation formats}
In Chinese font generation, target characters are often represented in two formats, bitmap font and vector font. As shown in Figure~\ref{Font Representation}, bitmap fonts typically encode each glyph as a fixed-resolution pixel grid, making it compatible with convolutional architectures and pixel-wise loss functions. They are relatively easy to process and have been widely used in early studies. However, since bitmap representations are resolution-dependent, they lack explicit structural information and are less flexible for structure-aware editing or scalable transformation.

In contrast, vector fonts mainly represent glyphs through geometric primitives (\textit{e.g.} strokes, contours or Bézier curves) and offer inherent resolution independence, which supports scalable rendering and better reflects the compositional structure of Chinese characters. Besides, its compact format facilitates efficient storage and enables flexible style manipulation. Nonetheless, vector-based font generation is more challenging, as it requires precise stroke modelling, structural consistency maintenance and specific supervision signals.

We summarize these two font representation formats in Table~\ref{bitmap_vs_vector}. In recent practice, bitmap-based methods are preferred due to their architectural simplicity and easy integration with mainstream generative models~\cite{157,165,160,168}, while vector-based approaches are increasingly explored for their advantages in high-resolution rendering and artistic expressiveness~\cite{151,152,163,169}.

\begin{table}[h]
\footnotesize
\centering
\caption{Comparisons between bitmap and vector font representation formats.}
\begin{spacing}{1.9}
\vspace{2mm}
\begin{tabular}{lll}
\toprule
Property & Bitmap font & Vector font \\
\midrule
Representation form & Pixel grid & Geometric primitives \\
Resolution dependency & Yes & No \\
Rendering flexibility & Low & High \\
Storage efficiency & Low & High \\
Strengths & Easy to process and good texture fidelity & Editable, scalable and structure-aware modelling \\
Weaknesses & Not scalable and lacks structural information & High generation complexity and limited datasets \\
\bottomrule
\end{tabular}
\end{spacing}
\label{bitmap_vs_vector}
\end{table}

\subsection{Public datasets}\label{datasetsection}
To facilitate Chinese font generation, a dataset containing a large number of Chinese characters and fonts is quite essential. Unfortunately, there is no widely-accepted standard dataset for this task so far. One pivotal reason is that many fonts are subject to copyright restrictions, which prevents researchers from openly sharing their own datasets. As a result, most studies rely on custom or private datasets, with only a few publicly available datasets released in the past years. In the following, we provide a synopsis and brief depiction of these accessible datasets.

\subsubsection{CASIA-HWDB}
The CASIA-HWDB datasets\footnote{\href{https://nlpr.ia.ac.cn/databases/handwriting/Download.html}{https://nlpr.ia.ac.cn/databases/handwriting/Download.html}} are built by the Institute of Automation of Chinese Academy of Sciences~\cite{34}, which can be used for character recognition, handwriting generation and other studies. The datasets include both online and offline data, which are divided into three isolated character datasets (DB1.0-1.2) and three handwritten text datasets (DB2.0-2.2). The isolated character datasets include about 3.9 million samples among 7,356 classes (including 7,185 Chinese characters and 171 symbols), while the handwritten text datasets comprise around 5,090 pages with 1.35 million character samples. Besides, each dataset is segmented and annotated at the character level.

\subsubsection{AGIS-Net}
The AGIS-Net dataset\footnote{\href{https://github.com/hologerry/AGIS-Net}{https://github.com/hologerry/AGIS-Net}}, created by the Wangxuan Institute of Computer Technology at Peking University, is a large-scale Chinese glyph images dataset covering a wide range of shape and texture styles for artistic style font generation~\cite{35}. This dataset is made up of 2,460 synthetic artistic fonts with 639 characters, as well as 35 artist-designed artistic fonts with 7,326 characters, which totals over 1.8 million images. 

\subsubsection{Fonts-100}
The Fonts-100 dataset\footnote{\href{https://github.com/liweileev/FET-GAN}{https://github.com/liweileev/FET-GAN}} is gathered by the College of Computer Science at Zhejiang University~\cite{38}. It encompasses 100 style fonts with 775 Chinese characters, 52 English letters, and 10 Arabic numerals. There are a total of 83,700 images, each of which is 320 $\times$ 320 in image dimension.

\subsubsection{FTransGAN}
The FTransGAN dataset\footnote{\href{https://github.com/ligoudaner377/font\_translator\_gan}{https://github.com/ligoudaner377/font\_translator\_gan}} is proposed by the Human Data Interaction Lab of Kyushu University~\cite{36}. It includes 847 gray-scale fonts, each with approximately 1,000 commonly-used Chinese characters and 52 Latin letters in the same style. All the images are uniformly resized to 64 $\times$ 64, which makes the dataset well-suited for cross-lingual font generation projects. 

\subsubsection{TE141K}
The TE141K dataset\footnote{\href{https://daooshee.github.io/TE141K/}{https://daooshee.github.io/TE141K/}}, introduced by the Wangxuan Institute of Computer Technology at Peking University, is an artistic text dataset with 141,081 text effect/glyph pairs in total~\cite{39}. It is composed of 152 professionally-designed text effects rendered on various glyphs like English letters, Chinese characters and Arabic numerals. In addition, each image has a resolution of 320 $\times$ 320.

\subsubsection{LSG-FCST}
The LSG-FCST dataset\footnote{\href{https://github.com/LYM1111/LSG-FCST}{https://github.com/LYM1111/LSG-FCST}} is put forward by Xi'an University of Technology~\cite{37}. It is separated into a training set and a testing set. For each of them, the training set includes 780 fonts, each containing 993 Chinese characters. The testing set is further classified into three kinds: UFUC (40 unseen fonts with 500 unseen characters), UFSC (29 unseen fonts with 993 seen characters) and SFUC (780 seen fonts with 29 unseen characters). Notably, there is no overlap between the training set and the testing set.

\begin{figure}[h]%
\centering
\includegraphics[width=0.77\textwidth]{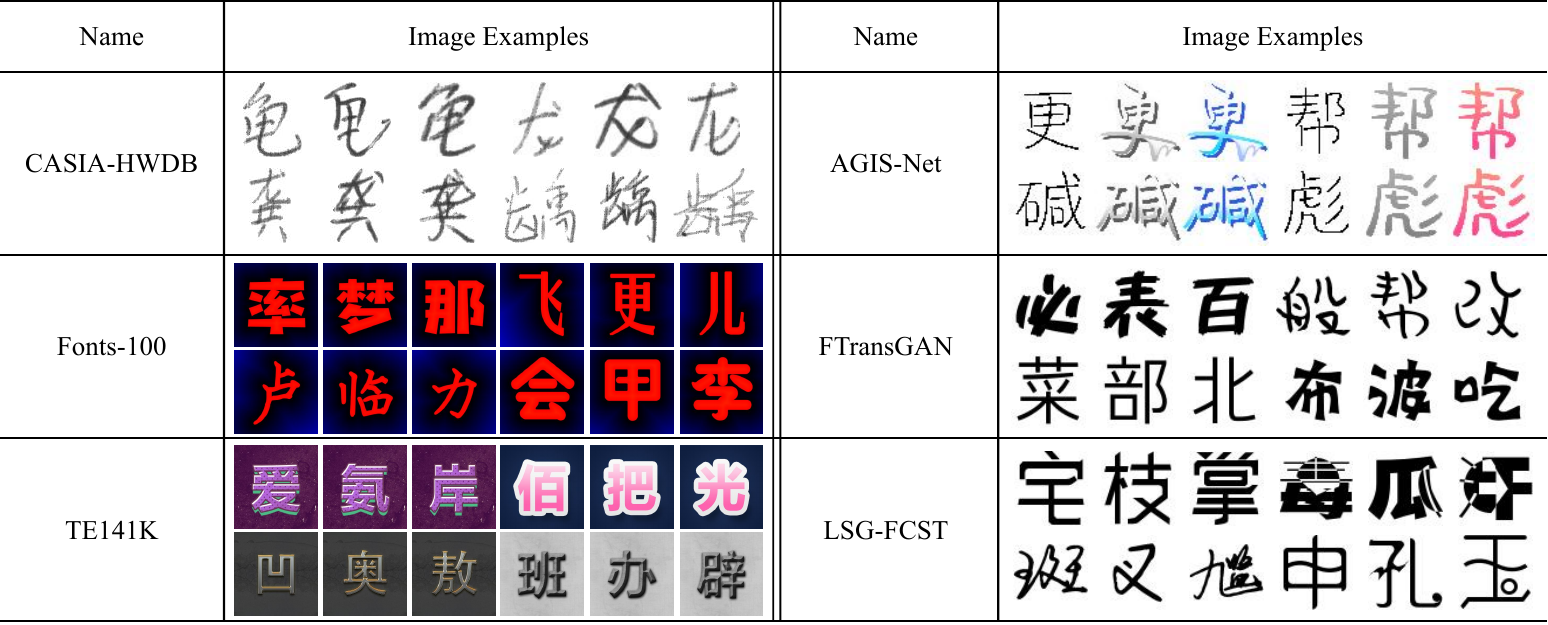}
\caption{Example images from the datasets in the present survey.}
\label{dataset examples}
\end{figure}

We sum up these above six datasets in Table~\ref{Datasetsummary} and present some sample images in Figure~\ref{dataset examples}, offering a useful reference for researchers to select the appropriate dataset for their individual assignments.

\begin{table}[h]
\centering
\footnotesize
\caption{Summary of publicly available datasets.}\label{Datasetsummary}
\begin{spacing}{1.9}
\vspace{2mm}
\begin{tabular}{lllll}
    \toprule
	Name & Year & Total images & Font \& character statistics \\
	\midrule
	\makecell[l]{CASIA-HWDB~\cite{34}}   & \makecell[l]{2011} &  \makecell[l]{About 5.25 million} & \makecell[l]{7,356 divergent classes of single Chinese characters and 5,090 pages of continuous ones}   \\
	\makecell[l]{AGIS-Net~\cite{35}}    & \makecell[l]{2019}  & \makecell[l]{Over 1.8 million}  & \makecell[l]{2,460 synthetic fonts with 639 characters and 35 artist-designed fonts with 7,326 glyphs}  \\
    \makecell[l]{Fonts-100~\cite{38}}  & \makecell[l]{2020} & \makecell[l]{83,700}  & \makecell[l]{100 style fonts with 775 Chinese characters, 52 English letters and 10 Arabic numerals}   \\
	\makecell[l]{FTransGAN~\cite{36}}   &  \makecell[l]{2021} & \makecell[l]{About 890,000} & \makecell[l]{847 gray-scale fonts with approximately 1,000 Chinese characters and 52 Latin letters}   \\
    \makecell[l]{TE141K~\cite{39}}   & \makecell[l]{2021} & \makecell[l]{141,081} & \makecell[l]{152 expertly-designed text effects rendered on Chinese characters, numerals and so on} \\
	\makecell[l]{LSG-FCST~\cite{37}}   &  \makecell[l]{2024} & \makecell[l]{About 1.2 million} & \makecell[l]{About 820 distinct fonts with approximately 1,493 commonly-used Chinese characters}    \\
	\bottomrule
\end{tabular}
\end{spacing}
\end{table}

\subsection{Evaluation metrics}
For the Chinese font generation task, evaluating the generated characters is a critical and intricate matter. At present, the quality evaluation metrics mainly include two classes: quantitative evaluation and qualitative evaluation.

\subsubsection{Quantitative evaluation}
Quantitative evaluation involves objectively assessing the quality of generated results using specific numerical indicators. Presently, a combination of multiple indicators is often leveraged to comprehensively measure the model's performance. In the following, we enumerate several ordinary evaluation metrics.

\textbf{MAE.} Mean Absolute Error (MAE), also known as L1 Loss, calculates the mean absolute difference between the pixel values of the generated characters and the ground truth~\cite{44}. It provides a clear assessment of the model’s performance, where a lower MAE value signifies that the generated characters more closely resemble the real ones. The calculation formula is as follows:

\begin{equation}\label{MAE}
\text{MAE} = \frac{\sum_{m=1}^M \sum_{n=1}^N \left| G(m,n) - R(m,n) \right|}{M \times N}
\end{equation}

\noindent{where $M$ and $N$ denote the length and width of the image, $G(m, n)$ and $R(m, n)$ represent the pixel value of the generated image and real image at the spatial position $(m,n)$, respectively.}

\textbf{MSE.} Mean Squared Error (MSE), also known as L2 Loss, captures the average squared difference between the generated characters and the ground truth~\cite{45}, which can be computed by:

\begin{equation}\label{MSE}
\text{MSE}=\frac{\sum_{m=1}^M\sum_{n=1}^N\left[G(m,n)-R(m,n)\right]^2}{M\times N}
\end{equation}

Following that, Root Mean Squared Error (RMSE) is another commonly-employed metric that builds on MSE by taking the square root of its value~\cite{41}. A lower RMSE value indicates greater similarity between images.

\textbf{PSNR.} Peak Signal-to-Noise Ratio (PSNR) is a widely-used measure of the signal reconstruction's quality in image compression~\cite{46}. Similar to RMSE, PSNR is also derived from MSE and expresses the evaluation result in decibels (dB), allowing it easier to compare across different images or compression algorithms. Let $L$ be the maximum pixel value of the image, PSNR can be formulated as: 

\begin{equation}\label{PSNR}
\text{PSNR}=10\ln\frac{L^2}{\text{MSE}}
\end{equation}

\textbf{SSIM.} Structural Similarity Image Metric (SSIM) primarily measures the similarity between the generated characters and the real ones, with a higher value indicating better performance~\cite{40}. Given a generated image $x$ and a real image $y$, SSIM assesses their luminance, contrast, and structural similarity, which are defined as follows:

\begin{equation}\label{luminance}
l\left(x,y\right)=\frac{2\mu_{x}\mu_{y}+C_{1}}{\mu_{x}^{2}+\mu_{y}^{2}+C_{1}}
\end{equation}

\begin{equation}\label{contrast}
c\left(x,y\right)=\frac{2\sigma_{x}\sigma_{y}+C_{2}}{\sigma_{x}^{2}+\sigma_{y}^{2}+C_{2}}
\end{equation}

\begin{equation}\label{structural}
s\left(x,y\right)=\frac{\sigma_{xy}+C_3}{\sigma_x\sigma_y+C_3}
\end{equation}

In Eqs. (\ref{luminance}), (\ref{contrast}) and (\ref{structural}), luminance is captured by the mean intensity $\mu_x$ and $\mu_y$, while contrast is estimated as the standard deviation $\sigma_x$ and $\sigma_y$. The structure similarity is indicated by the correlation coefficient $\sigma_{xy}$ between $x$ and $y$. In order to enhance numerical stability, a small constant is added to both the denominator and numerator of the three equations. Then, the SSIM index is obtained via combining these three components, which is calculated by:

\begin{equation}\label{SSIM}
\mathrm{SSIM}(x,y)=[l(x,y)]^\alpha\cdot[c(x,y)]^\beta\cdot[s(x,y)]^\gamma 
\end{equation}

\noindent{where $\alpha$, $\beta$, and $\gamma$ are three hyper-parameters utilized to adjust the significance of pertinent information. Through taking high-level vision to grasp image quality, SSIM can circumvent the complexity of low-level visual modeling.}

\textbf{FID.} Frechet Inception Distance (FID) is a popular metric to calculate the Frechet distance (also called Wasserstein-2 distance) between the feature vectors of the generated characters and real ones, with these features being extracted by a pre-trained Inception network~\cite{43}. This distance is computed based on the mean and covariance of the feature vectors, and the concrete formula is as follows:

\begin{equation}\label{FID}
\mathrm F\mathrm I\mathrm D=\|\mu_r-\mu_g\|_2^2+\mathrm T\mathrm r(\Sigma_r+\Sigma_g-2(\Sigma_r\Sigma_g)^{1/2})
\end{equation}

In Eq.~(\ref{FID}), $\mu_r$ and $\mu_g$ represent the mean feature vectors of the real and generated images, while $\sigma_r$ and $\sigma_g$ denote their respective covariance matrices. $T_r$ refers to the trace of a matric, which is the sum of its diagonal elements. A lower FID score indicates that the generated images are more similar to the real ones, both in terms of individual features and their overall distribution.

\textbf{LPIPS.} Learned Perceptual Image Patch Similarity (LPIPS) is an advanced metric designed to evaluate the perceptual similarity between images~\cite{42}. Unlike traditional pixel-level metrics, LPIPS takes advantage of the deep neural network pre-trained on human perceptual data to compare differences in higher-level feature space, which enables it more aligned with human visual perception. The formula for LPIPS is defined by: 

\begin{equation}\label{LPIPS}
\mathrm{LPIPS}(x,y)=\sum_l\frac{1}{H_lW_l}\sum_{h,w}\left\|\phi_l(x)_{hw}-\phi_l(y)_{hw}\right\|_2^2
\end{equation}

\noindent{where $x$ and $y$ are the two images being compared, $\phi_l(x)$ and $\phi_l(y)$ denote the feature maps extracted from a specific network layer $l$. The dimensions of these feature maps are represented by $H_l$ and $W_l$, with $h$ and $w$ referring to the spatial coordinates within the feature maps.}

\subsubsection{Qualitative evaluation}
Human perception of beauty cannot be easily reflected through quantitative metrics alone. Therefore, some studies~\cite{130,146,160} also employ qualitative evaluation strategies, such as visual comparison analysis and user study, to assess the ability of models in replicating details and overall styles.

\textbf{Visual comparison analysis.} Visual comparison analysis refers to the process of comparing the generated Chinese characters with the corresponding ground truth to assess their similarity in visual effects, such as glyph structure, stroke details and style consistency. It allows researchers to intuitively examine not only the accuracy of individual strokes but also the harmony and balance of the overall character structure. 

\textbf{User study.} User study is another widely-used approach for directly evaluating the generation quality. In such studies, participants are shown images produced by different methods and asked to visually assess the quality of each character, typically by selecting the most preferred one. The aggregated user preferences offer intuitive and valuable insights into the perceived realism of the generated characters, serving as important feedbacks for model refinements.

\section{Many-shot Chinese font generation approaches}\label{Many-shot Chinese font generation approaches}
\begin{figure}[h]%
\centering
\includegraphics[width=0.7\textwidth]{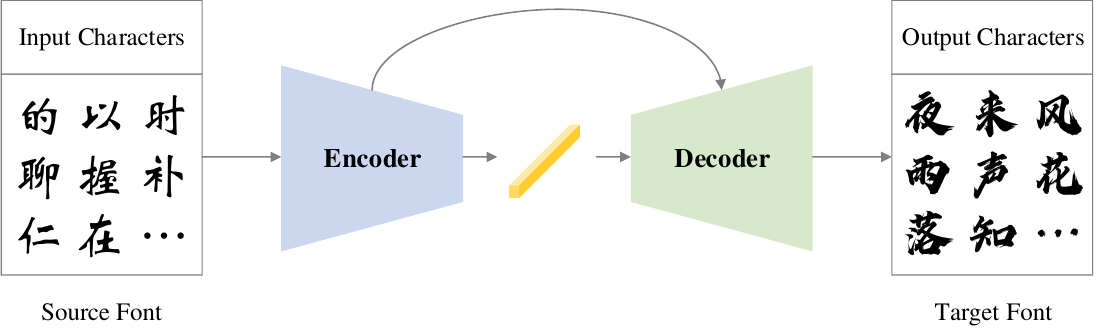}
\caption{An overview diagram of many-shot Chinese font generation methods.}\label{Overview of many-shot generation methods}
\end{figure}

Since font generation can be formulated as mapping font images from a source domain to a target domain, preliminary approaches often employ Image-to-Image (I2I) methods to learn the mapping function between different fonts, which is mainly designed in the form of an encoder-decoder framework. An overview diagram of such approaches is shown in Figure~\ref{Overview of many-shot generation methods}. In order to guarantee the precision of the generated results, the mapping function needs to be fine-tuned on hundreds of reference samples (also known as many-shot) to generate unseen font images~\cite{47}. According to the characteristics of the training data, we categorize many-shot font generation methods into two types: paired-data-based methods and unpaired-data-based methods.

\subsection{Paired-data-based methods}
Paired-data-based methods generally use numerous pairs of source font and target font images to learn a direct mapping between them. Tian~\cite{48} first puts forward a top-down CNN structure called Rewrite, which consists of multiple convolutional layers with batch normalization and is trained by minimizing the pixel-wise MAE. However, when there is a significant stylistic discrepancy between the source font and the target font, the generated results would be very blurry or even illegible. Then, based upon the classical I2I translation framework pix2pix~\cite{51}, zi2zi~\cite{50} utilizes one-hot style label vectors to optimize the font generation model in a supervised manner, which only results in good synthesizing performance on some specific font styles. DC-Font~\cite{55} expands zi2zi by introducing an additional style classifier to get better style representations and then synthesizes the target character via a generative network, but for characters written in highly-cursive styles, the output's quality remains unsatisfactory. Besides, HGAN~\cite{52} further extends the zi2zi model via a transfer network and a hierarchical adversarial discriminator. Xiao~\cite{24} adds a bridge structure between the encoder and decoder of VAE to promote the information transfer within the model. Whereas, the generated characters still suffer from poor clarity. Chen et al.~\cite{10} adopt a new feature extraction method (including clustering, segmentation, denoising and so on) to get a better feature set, but it only focuses on image-level features while ignoring the minutia information. AEGG~\cite{56} uses an extra auto-encoder network to provide supervisory information during the training process. Nevertheless, it is limited to single-class character generation.

To surmount this problem, PEGAN~\cite{59} sets up a multi-scale pyramid of refinement information into U-Net structure~\cite{60} and links it to the corresponding feature maps in the decoder. This ingenious design can make the generator preserve more precise information. Likewise, Zhang et al.~\cite{69} put forward a multi-scale GAN, where the small-scale GAN generates character outline and the large-scale one supplements fine details, but for the complex Chinese characters with dense strokes, the model struggles to generate complete semantic glyphs. Zhang~\cite{61} raises an end-to-end generation model based on the residual-dense connection and integrates a Hybrid Dilated Convolution (HDC)~\cite{80} structure to enhance feature transfer across different scales. Nonetheless, the generated images are of fixed size, which confines the model's adaptability in different scenarios. SAFont~\cite{62} utilizes self-attention mechanisms to model the relationships between strokes and defines an edge loss to produce sharper stroke edges. Grounded in WGAN-GP~\cite{49}, Liang~\cite{31} introduces a dual discriminator to independently judge the font content and style. Afterwards, FTFNet~\cite{72} captures the detailed structural features through local residual learning and global feature fusion. Whereas, since the model explicitly learns the transformation from a particular source style to a given target style, it lacks the universalization capability to new styles. 

\begin{figure}[h]
    \centering
    \includegraphics[width=0.72\linewidth]{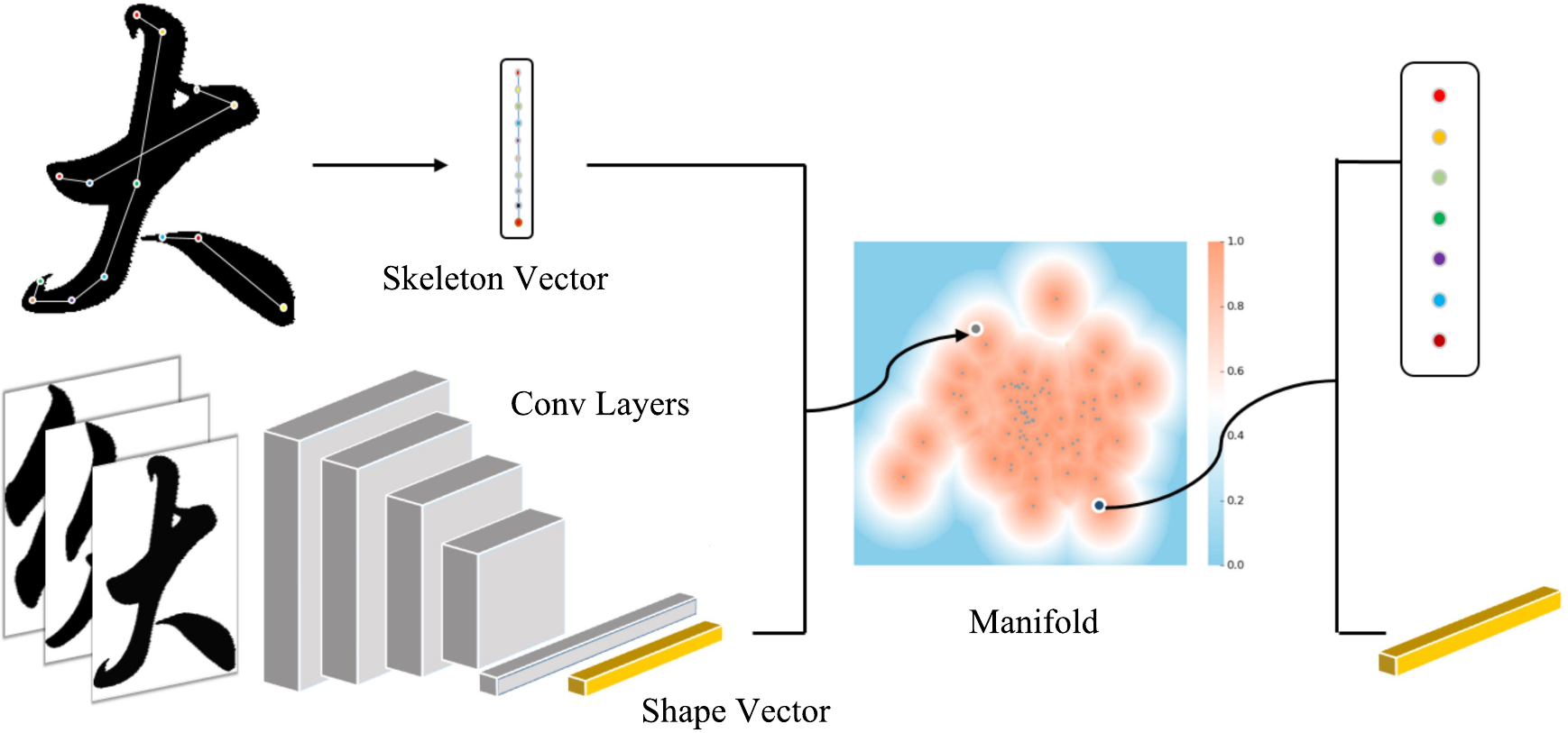}
    \caption{An overview of the manifold building procedure in Guo et al~\cite{58}.}
    \label{manifold}
\end{figure}

\begin{figure}[htbp]
    \centering
    \includegraphics[width=0.56\linewidth]{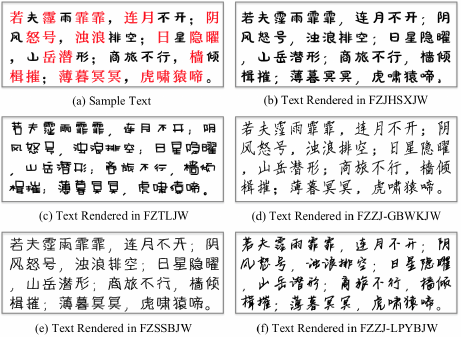}
    \caption{Texts rendered in five different font libraries generated by SCFont~\cite{63}. The synthesized characters in (b-f) are marked in red in (a).}
    \label{SCFont Results}
\end{figure}

Apart from the above approaches, some methods also take advantage of stroke-level information to further polish up the generation process. Lian et al.~\cite{53} develop a handy system for synthesizing personalized handwriting, which decomposes the overall handwriting style into the stroke shape style and stroke layer style. Later, inspired by CGAN~\cite{26}, Liu et al.~\cite{12} propose a novel generative adversarial network called MSMC-CGAN, which combines both global information and local details as the condition of CGAN. After that, drawing from manifold learning, Guo et al.~\cite{58} extract feature vectors to represent each Chinese character's skeleton and outline shape, and map them to a low-dimensional manifold by virtue of a dimension reduction technique (as shown in Figure~\ref{manifold}). Nevertheless, the jointly changing process occasionally fails due to the inevitable reconstruction loss from the low-dimensional space to the high-dimensional space. SCFont~\cite{63} separates the font generation procedure into two steps, writing trajectories synthesis and font style rendering, which ensures both structural accuracy and style consistency in the generated characters. Figure~\ref{SCFont Results} presents some representative results. Gao et al.~\cite{64} exploit a data-driven system called ChinFont. It voluntarily divides the input glyphs into vectorized components and strokes, and also includes a layout prediction module to learn the structure as well as layout information of Chinese characters. CalliGAN~\cite{66} first segments Chinese characters into dictionary sequences and then encodes them through a Recurrent Neural Network (RNN), which offers richer low-level structure information. Instead of directly generating character images, SSNet~\cite{67} synthesizes target typographies by utilizing the disentangled stroke features from a structure module along with the pre-trained semantic features from a semantic module. Whereas, since SSNet adopts a two-phrase training strategy, the training cost dramatically increases. Wen et al.~\cite{70} propose a CNN-based model with three new techniques: collaborative stroke refinement, online zoom-augmentation and adaptive pre-deformation to standardize and align the characters. Wang et al.~\cite{71} introduce the prior information of stroke semantics to mitigate incorrect stroke generation, but their method is impactful for standard printed Chinese fonts alone. CVFont~\cite{75} leverages a deep object detection framework to infer the layout of Chinese characters, which is then applied to assemble selected vector components and generate glyphs for unseen characters. Even so, it still struggles with characters that have duplicated or small components and complex layouts. Moreover, the performance heavily depends on the accuracy of the component extraction algorithm. Liu et al.~\cite{77} bring in a Deformation Attention Skip-Connection (DASC) module that learns the offsets and weights via cross-channel interaction between features. Meanwhile, through mapping the stroke-type semantic categories to the encoder, the model can better utilize stroke category information during the generation process.

We provide a concise summary of these paired-data-based generation methods in Table~\ref{pairdatabasedsummary}. On the positive side, these methods are good at encapsulating the accurate relationships between the source font and the target font. Nonetheless, they rely heavily on paired data, which would be time-consuming and costly to gather at scale, and even in some special circumstances, paired images may not be obtainable at all. On top of that, their generalization ability is confined, making them challenging to handle unseen font styles or Chinese characters. These shortcomings restrict the broader applications of such methods. 
 
\begin{table}[!h]
\centering
\footnotesize
\caption{A summary of presented paired-data-based methods.}\label{pairdatabasedsummary}
\begin{spacing}{1.9}
\vspace{2mm}
\begin{tabular}{lll}
    \toprule
    Literature & Year & Brief description  \\
    \midrule
    \makecell[l]{Rewrite~\cite{48}}   & \makecell[l]{2016} &  \makecell[l]{Propose a CNN with multiple convolutional layers and train it by minimizing the pixel-wise MAE}   \\
    \makecell[l]{Lian et al.~\cite{53}}   & \makecell[l]{2016} &  \makecell[l]{Decompose the overall handwriting style into stroke shape style and stroke layer style respectively}   \\
    \makecell[l]{zi2zi~\cite{50}}    & \makecell[l]{2017}  &  \makecell[l]{Utilize one-hot style label vectors to optimize the font generation procedure in a supervised mode}   \\
    \makecell[l]{DCFont~\cite{55}}   &  \makecell[l]{2017} &  \makecell[l]{Introduce an extra style classifier and then synthesize the target character via a generative network}    \\
    \makecell[l]{AEGG~\cite{56}}   &  \makecell[l]{2017} &  \makecell[l]{Bring in an additional auto-encoder network to offer supervisory information during training stage}    \\
    \makecell[l]{Xiao et al.~\cite{24}}   &  \makecell[l]{2018} &  \makecell[l]{Insert a bridge structure between the encoder and decoder of VAE to promote information transfer}    \\
    \makecell[l]{PEGAN~\cite{59}}   &  \makecell[l]{2018} &  \makecell[l]{Add a multi-scale pyramid into U-Net and link it to the corresponding feature maps in the decoder}    \\
    \makecell[l]{HGAN~\cite{52}}   &  \makecell[l]{2018} &  \makecell[l]{Extend the zi2zi architecture through a transfer network and a hierarchical adversarial discriminator}    \\
    \makecell[l]{Guo et al.~\cite{58}}   &  \makecell[l]{2018} &  \makecell[l]{Extract feature vectors to represent skeleton and outline and map them to a low dimension manifold}    \\
    \makecell[l]{Zhang~\cite{61}}  & \makecell[l]{2019} & \makecell[l]{Raise an end-to-end generation model based upon the residual-dense connection and HDC structure}   \\
    \makecell[l]{SAFont~\cite{62}}  & \makecell[l]{2019} & \makecell[l]{Use self-attention to model the strokes' relationships and define an edge loss to produce sharp edges}   \\
    \makecell[l]{SCFont~\cite{63}}  & \makecell[l]{2019} & \makecell[l]{Decompose the font generation procedure into writing trajectories synthesis and font style rendering}   \\
    \makecell[l]{MSMC-CGAN~\cite{12}}  & \makecell[l]{2019} & \makecell[l]{Combine both global feature information and local details together as the condition input of CGAN}   \\
    \makecell[l]{ChinFont~\cite{64}}  & \makecell[l]{2019} & \makecell[l]{Separate glyphs into components, and leverage a prediction module to learn the structure and layout}   \\
    \makecell[l]{SSNet~\cite{67}}  & \makecell[l]{2020} & \makecell[l]{Deploy disentangled stroke feature and pre-trained semantic features to generate the target characters}   \\
    \makecell[l]{CalliGAN~\cite{66}}  & \makecell[l]{2020} & \makecell[l]{Disassemble Chinese characters into dictionary sequences and encode them through an RNN model}   \\
    \makecell[l]{Liang~\cite{31}}  & \makecell[l]{2021} & \makecell[l]{Present a dual discriminator to separately judge the font content and font style based on WGAN-GP}   \\
    \makecell[l]{FTFNet~\cite{72}}  & \makecell[l]{2021} & \makecell[l]{Capture the detail features of font style via a local residual learning and global feature fusion module}   \\
    \makecell[l]{Wen et al.~\cite{70}}  & \makecell[l]{2021} & \makecell[l]{Raise a CNN-based model with stroke refinement, zoom-augmentation and adaptive pre-deformation}   \\
    \makecell[l]{Zhang et al.~\cite{69}}  & \makecell[l]{2022} & \makecell[l]{Design a multi-scale GAN where small-scale one produce outlines and large-scale one adds minutiae}   \\
    \makecell[l]{Wang et al.~\cite{71}}  & \makecell[l]{2022} & \makecell[l]{Integrate prior stroke semantics information and a skip connection module into the network structure}   \\
    \makecell[l]{Chen et al.~\cite{10}}  & \makecell[l]{2022} & \makecell[l]{Take advantage of a novel feature extraction method so as to acquire a better feature set of characters}   \\
    \makecell[l]{CVFont~\cite{75}}  & \makecell[l]{2022} & \makecell[l]{Apply an object detection framework to infer layout and assemble components for unseen characters}   \\
    \makecell[l]{Liu et al.~\cite{77}}  & \makecell[l]{2024} & \makecell[l]{Utilize a deformation-attention skip-connection module to learn offsets and weights between features}   \\
    \bottomrule
\end{tabular}
\end{spacing}
\end{table}

\subsection{Unpaired-data-based methods}
\begin{figure}[h]%
\centering
\includegraphics[width=0.25\textwidth]{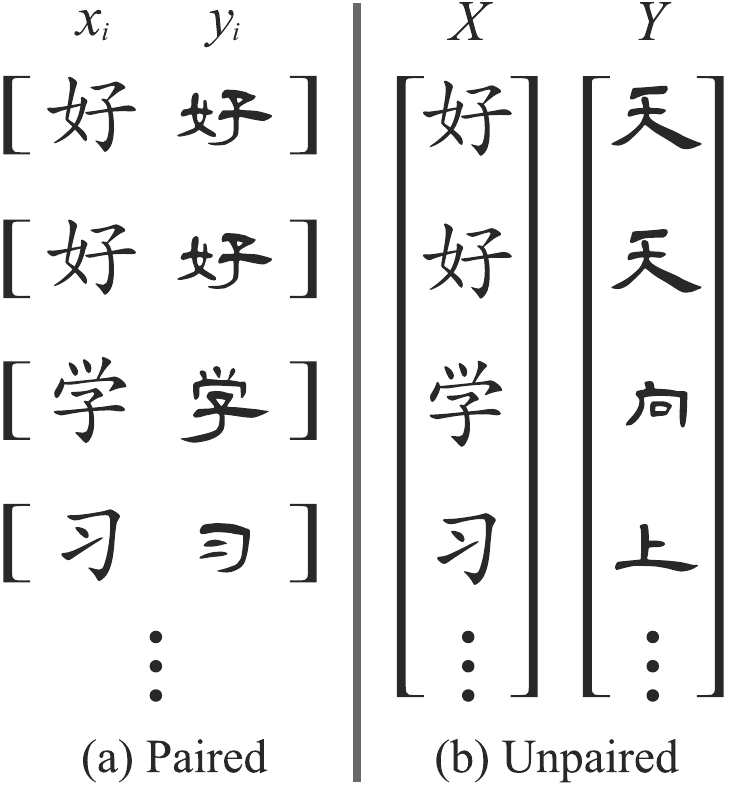}
\caption{Comparison between paired training data and unpaired training data~\cite{88}. (a) In the paired setting, each source sample $x_i$ is aligned with its corresponding target annotation $y_i$. (b) In the unpaired setting, the source domain $X$ and target domain $Y$ include independent samples.}\label{Paired data vs unpaired data}
\end{figure}

\begin{figure}[h]%
\centering
\includegraphics[width=0.89\textwidth]{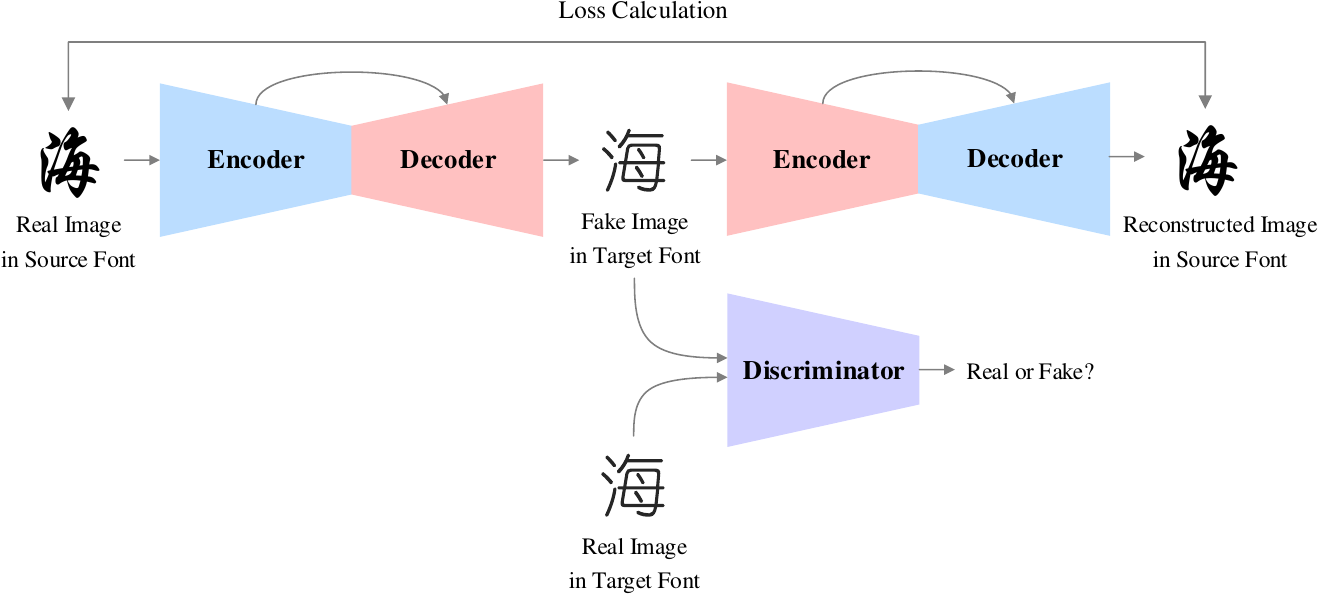}
\caption{General architecture of the unpaired-data-based Chinese font generation methods.}\label{General architecture of an unpaired-data based Chinese font generation method}
\end{figure}

Unpaired-data-based methods for many-shot Chinese font generation are mainly built on the classical CycleGAN architecture~\cite{27}. They leverage the cycle consistency mechanism to transfer font styles between the source and target domains without the need for paired examples. As illustrated in Figure~\ref{Paired data vs unpaired data}, we differentiate between paired and unpaired training data to clarify the data assumptions of these methods. Additionally, the loss calculation between the reconstructed image and the real image is included to ensure accurate transformation as well. The general framework of these approaches is displayed in Figure~\ref{General architecture of an unpaired-data based Chinese font generation method}.

\begin{figure}[t]%
\centering
\includegraphics[width=0.7\textwidth]{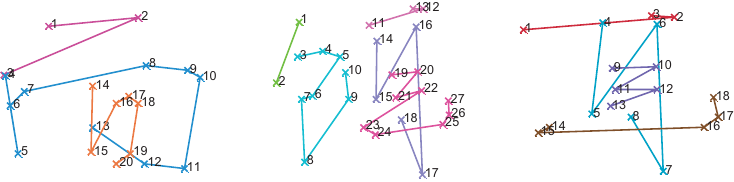}
\caption{Illustration of three Chinese characters in the sequential data format~\cite{83}.}\label{sequential data format}
\end{figure}

Chang et al.~\cite{81} put forward a generation method called HCCG-CycleGAN. It uses DenseNet~\cite{109} as part of the generator, which achieves the mapping function from printing typeface to personalized writing style. Zhang et al.~\cite{82} propose a framework that leverages RNN as both a discriminative model for recognizing Chinese characters along with a generative model for drawing them. This approach directly handles the sequential structure without the requirement for domain-specific knowledge, but it can only be applied to online handwritten Chinese characters. Following that, FontRNN~\cite{83} models the Chinese characters as sequences of points (as shown in Figure~\ref{sequential data format}) and introduces a RNN model with a monotonic attention mechanism. Despite that, FontRNN also has some drawbacks. It is plagued by the absence of strokes owing to imperfections in the reference dataset. What's more, since FontRNN predicts each timestep's output based on the previous one, the total error would accumulate significantly over long sequences and thereby degrade the final result. FontGAN~\cite{84} combines the stylization, de-stylization, and texture transfer of Chinese characters into a unified framework. Meanwhile, to alleviate problems posed by multi-domain transformations, it also incorporates a style consistency module and a content prior module to ensure more coherent character generation. Zhang~\cite{86} raises a model called OFM-CycleGAN, which employs the Optimized Feature Matching (OFM) algorithm in the forward as well as backward mapping processes of CycleGAN to provide more statistical information for the model. MTfontGAN~\cite{89} leverages a multitask adversarial learning approach that learns distinct mappings among multiple Chinese fonts using just a single generator and multiple discriminators. MSHCC-GAN~\cite{90} utilizes a novel generator module called Gttention. It can differentiate between various styles during the generation process, so as to realize the one-to-many conversion across different fonts. Informed by transfer learning, Zhang~\cite{91} employs a two-stage training method, where a large model is trained to convert printed fonts into handwritten ones in the first stage and then the model is fine-tuned with a small set of handwritten Chinese character samples in the second stage. ChiroGAN~\cite{92} capitalizes on a three-stage architecture for multi-chirography image translation, which is made up of skeleton extraction, skeleton transformation and stroke rendering. Yet, it relies on the effects of corrosion and expansion algorithms, so that it often crashes on complex characters or irregular styles.

\begin{figure}[t]%
\centering
\includegraphics[width=\textwidth]{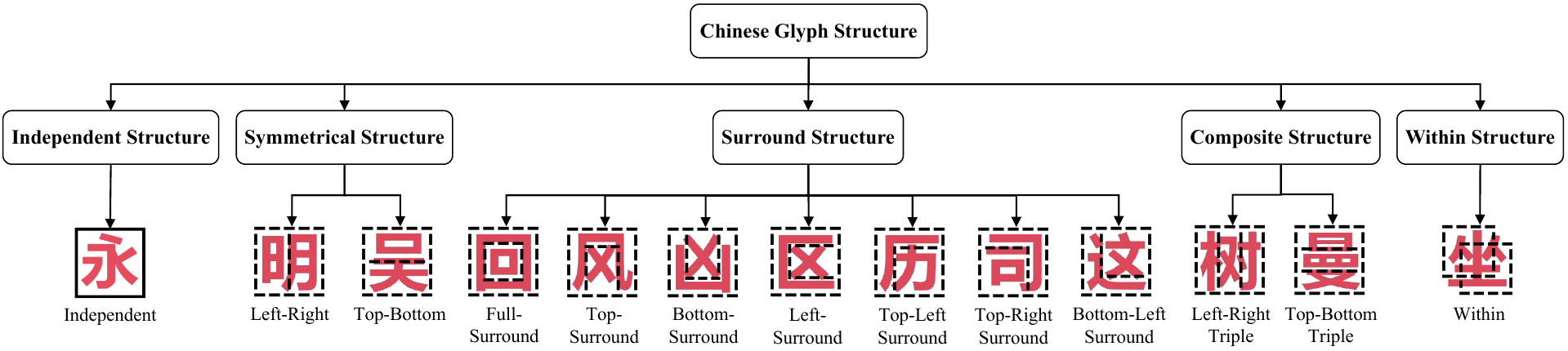}
\caption{Common spatial structures of Chinese characters.}
\label{Common spatial structures of Chinese characters}
\end{figure}

Building upon the StarGAN structure~\cite{93}, Lin et al.~\cite{94} put forth an unsupervised style transfer model. Within the model, the generator is strengthened with a DenseNet~\cite{109} placed between the encoder and decoder, while the discriminator is refined via inserting additional convolutional branches. Zeng et al.~\cite{97} integrate a certain diversity regularizer to the total training loss, so that the generator can output high-quality characters with better variation. Zhou et al.~\cite{88} insert an extra content supplement network within the style transfer model. It can efficiently capture the low-dimensional information to boost the final output. CS-GAN~\cite{29} associates distribution transform, reparameterization trick, and sampling features together to convert feature maps obtained from the source domain to the target domain. However, it concentrates on the font's overall structure while ignoring the finer stroke details. Mu~\cite{54} incorporates a dense connection network and U-Net into the generator, so as to strengthen the retention of stroke structure information of input Chinese characters. Later, inspired by the thirteen spatial structures of Chinese characters (as shown in Figure~\ref{Common spatial structures of Chinese characters}), RCN~\cite{95} views each character as a composition of radicals rather than a single unit. It utilizes a self-recurrent network as an encoder to integrate the radicals into a vector, which is then combined with the encoder's output for further processing. Hassan et al.~\cite{98} put forward a method that uses a single generator to conditionally produce various font family styles. StrokeGAN~\cite{30} employs one-bit stroke encoding to capture the key mode information of Chinese characters, but it only uses the primal stroke and neglects the component labels. Liu et al.~\cite{96} decouple character images into style representation and content representation, respectively. To better encapsulate the style information, a style consistency module along with a content prior module are also leveraged to model the style representations into different prior distributions. Yet, it performs poorly on the font de-stylization inference with novel cursive or intricate styles. 

\begin{figure}[h]%
\centering
\includegraphics[width=0.6\textwidth]{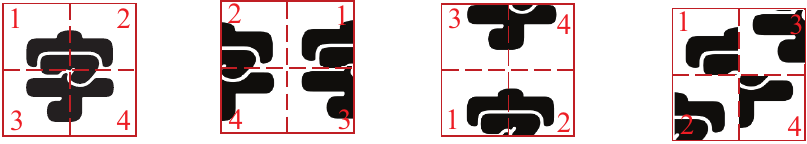}
\caption{Square-block geometric transformations.}
\label{Tianzige}
\end{figure}

Subsequently, motivated by the observation that the skeleton embodies both local and global structure information of Chinese characters, SGCE-Font~\cite{101} implements the skeleton information into the generator via the channel expansion operation without adding any extra model complexity. Zeng et al.~\cite{102} make use of a squared-block transformation-based self-supervised method to guide the model network to extract features with higher quality, as illustrated in Figure~\ref{Tianzige}. The proposed squared-block geometric transformation does not demand any modification of existing model networks and additional human labour cost, which can be lightly embedded in many existing models. Inspired by DDPM~\cite{33}, Liao et al.~\cite{103} present a system called Calliffusion, which leverages external information, such as Chinese text descriptions of character, script, and style, as the control condition. Noticeably, Callifusion does not necessitate the use of any images during inference. Despite this, the generated results sometimes might be afflicted by the stroke errors (including missing or redundant strokes). RC-GAN~\cite{105} enforces the radical sequence of the Chinese character to be correctly decoded from the generated results. To fulfill this, a Gated Recurrent Unit (GRU)-based radical learning network~\cite{166} is introduced to capture the radical components, while a radical consistency loss is also applied to optimize the training of this module.

We present a brief resume of these unpaired-data-based generation methods in Table~\ref{unpairdatabasedsummary}. Compared with paired-data-based methods, these approaches reduce both the cost and effort required to collect large volumes of labelled data. Along with that, they also enable more flexible and dynamic transformations between the source font and the target font. Nevertheless, since unpaired data is used during training, the generated characters may not be able to completely match the original characters in terms of semantic content, leading to potential inconsistencies in the structure details like missing or redundant construction.

\begin{table}[!h]
\centering
\footnotesize
\caption{A summary of presented unpaired-data-based methods.}\label{unpairdatabasedsummary}
\begin{spacing}{1.9}
\vspace{2mm}
\begin{tabular}{lll}
    \toprule
    Literature & Year & Brief description  \\
    \midrule
    \makecell[l]{HCCG-CycleGAN~\cite{81}}   &  \makecell[l]{2018} &  \makecell[l]{Take advantage of DenseNet as a portion of the CycleGAN generator to enhance the outputs' quality}    \\
    \makecell[l]{Zhang et al.~\cite{82}}   &  \makecell[l]{2018} &  \makecell[l]{Exploit RNN as both a discriminative and a generative model for recognizing and drawing characters}    \\
    \makecell[l]{FontRNN~\cite{83}}  & \makecell[l]{2019} & \makecell[l]{Model the characters as sequences of points and introduce an RNN model with a monotonic attention}   \\
    \makecell[l]{FontGAN~\cite{84}}  & \makecell[l]{2019} & \makecell[l]{Combine stylization, de-stylization and texture transfer of Chinese characters into a unified framework}   \\
    \makecell[l]{OFM-CycleGAN~\cite{86}}  & \makecell[l]{2019} & \makecell[l]{Employ the OFM algorithm in the forward as well as backward mapping processes of the CycleGAN}   \\
    \makecell[l]{MTfontGAN~\cite{89}}  & \makecell[l]{2020} & \makecell[l]{Make use of a multitask adversarial learning approach to learn distinct mappings among Chinese fonts}   \\
    \makecell[l]{MSHCC-GAN~\cite{90}}  & \makecell[l]{2020} & \makecell[l]{Utilize a novel generator module Gttention to distinguish between various styles during the generation}   \\
    \makecell[l]{Zhang~\cite{91}}  & \makecell[l]{2020} & \makecell[l]{Use a two-stage training method: train on regular font transfer and fine-tune with handwritten samples}   \\
    \makecell[l]{ChiroGAN~\cite{92}}  & \makecell[l]{2020} & \makecell[l]{Apply a three-stage GAN model for skeleton extraction, skeleton transformation, and stroke rendering}   \\
    \makecell[l]{Lin et al.~\cite{94}}  & \makecell[l]{2020} & \makecell[l]{Improve the generator with DenseNet and refine discriminator through adding convolutional branches}   \\
    \makecell[l]{Zeng et al.~\cite{97}}  & \makecell[l]{2021} & \makecell[l]{Integrate a certain diversity regularizer into the training loss so as to generate higher-quality characters} \\
    \makecell[l]{Zhou et al.~\cite{88}}  & \makecell[l]{2021} & \makecell[l]{Insert a content supplement network within the style transfer model to capture the style strokes' details}   \\
    \makecell[l]{CS-GAN~\cite{29}}  & \makecell[l]{2021} & \makecell[l]{Associate distribution transform, reparameterization trick and feature sampling to ensure style transfer}  \\
    \makecell[l]{Mu~\cite{54}}  & \makecell[l]{2021} & \makecell[l]{Incorporate a dense connection network and U-Net to strengthen the reservation of stroke information}   \\
    \makecell[l]{RCN~\cite{95}}  & \makecell[l]{2021} & \makecell[l]{Set up a self-recurrent network to integrate radicals in a vector and concat it with the encoder's output }   \\
    \makecell[l]{Hassan et al.~\cite{98}}  & \makecell[l]{2021} & \makecell[l]{Make use of only one generator to conditionally produce a number of font styles to form a font family}   \\
    \makecell[l]{StrokeGAN~\cite{30}}  & \makecell[l]{2021} & \makecell[l]{Leverage one-bit stroke encoding strategy to encapsulate key mode information of Chinese characters}   \\
    \makecell[l]{Liu et al.~\cite{96}}  & \makecell[l]{2022} & \makecell[l]{Decouple Chinese character images into independent style representations and content representations}   \\
    \makecell[l]{SGCE-Font~\cite{101}}  & \makecell[l]{2022} & \makecell[l]{Append the skeleton information into the generator structure through the channel expansion operation}   \\
    \makecell[l]{Zeng et al.~\cite{102}}  & \makecell[l]{2022} & \makecell[l]{Put forward a squared-block transformation-based self-supervised method to guide the model training}  \\
    \makecell[l]{Callifusion~\cite{103}}  & \makecell[l]{2023} & \makecell[l]{Adopt the outer information like text descriptions of character, script and style, as the control condition}   \\
    \makecell[l]{RC-GAN~\cite{105}}  & \makecell[l]{2024} & \makecell[l]{Implement a GRU-based radical learning structure to extract the radical components within characters}   \\
    \bottomrule
\end{tabular}
\end{spacing}
\end{table}

In Table~\ref{summary of many shot methods}, we comb the two classes of many-shot Chinese font generation methods analyzed above. Although these methods can generate promising font images, gathering large quantities of reference samples with a coherent style is a troublesome and expensive task. Furthermore, balancing the complexity of the model with the availability of training data is critical to realizing better performance. 

\begin{table}[htbp]
\centering
\footnotesize
\caption{Classification of many-shot Chinese font generation approaches.}\label{summary of many shot methods}
\begin{spacing}{1.9}
\vspace{2mm}
\begin{tabular}{llll}
    \toprule
    Type of method & Simplistic depiction & Strengths & Limitations   \\
    \midrule
    \makecell[l]{Paired-data-based}   &  \makecell[l]{Use numerous pairs of font images \\ to learn a direct mapping function} &  \makecell[l]{Excel in capturing the relationships \\ between the source and target font}  & \makecell[l]{Rely on the paired data and the \\generalization ability is limited}   \\ 
    \makecell[l]{Unpaired-data-based}   &  \makecell[l]{Leverage the cycle consistency to \\ transfer between different fonts} &  \makecell[l]{Reduce the cost to make paired data \\ and enable more flexible mutations}  & \makecell[l]{The generated results may not \\fully match the original ones}  \\
    \bottomrule
\end{tabular}
\end{spacing}
\end{table}

\section{Few-shot Chinese font generation approaches}\label{Few-shot Chinese font generation approaches}

\begin{figure}[h]%
\centering
\includegraphics[width=0.8\textwidth]{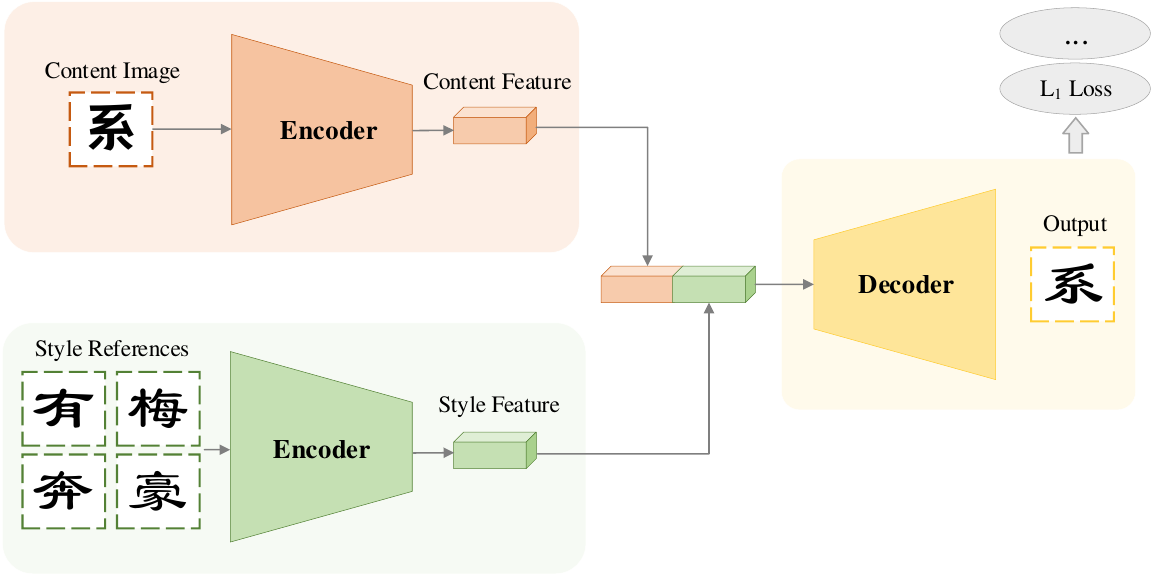}
\caption{A logic diagram of few-shot Chinese font generation methods.}\label{Overview of few-shot generation methods}
\end{figure}

To relieve the weaknesses of many-shot Chinese font generation models, few-shot font generation has become a hot topic in recent years, which intends to transfer the font style from a source domain to a target domain using only a few reference images. As of now, most few-shot font generation methods try to disentangle the content and style features from the given characters, respectively. A logic diagram of such approaches is described in Figure~\ref{Overview of few-shot generation methods}. Based on different kinds of feature representations, we can divide the prevailing generation methods into two categories: universal-feature-based methods and structural-feature-based methods. 

\subsection{Universal-feature-based methods}
\begin{figure}[t]
    \centering
    \includegraphics[width=0.87\linewidth]{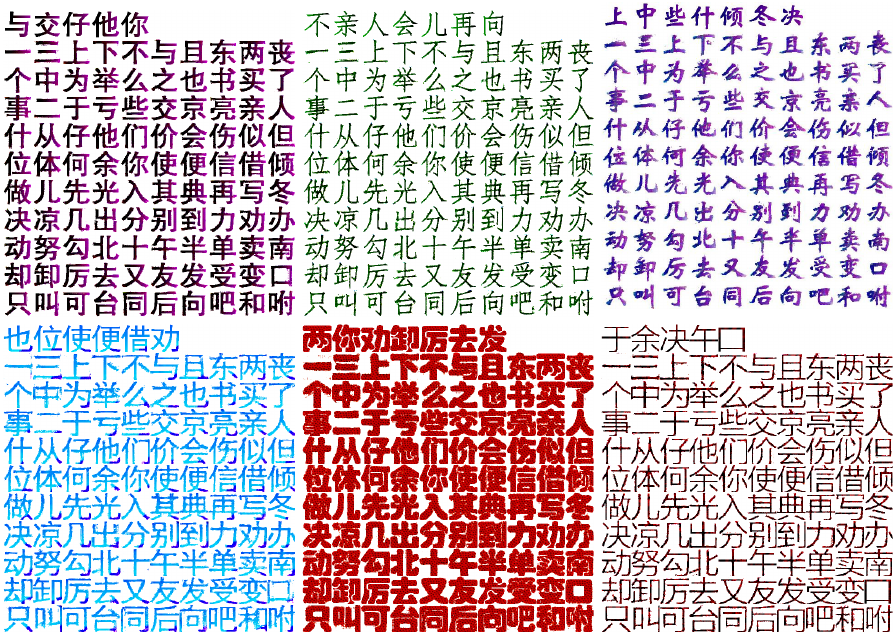}
    \caption{Some generation results in~\cite{122}. The first row of each style shows the reference images and the other characters are synthesized results.}
    \label{Zhuanna}
\end{figure}

Universal-feature-based methods predominantly synthesize new glyphs by merging the style features encoded from the reference set and the content features encoded from the source glyph straightly. To begin with, EMD~\cite{110} extracts the style and content features from a set of style-reference images (sharing the same style but different contents) and content images (sharing the same content but different styles), respectively. These features are then blended in a bilinear mixer network to generate final results. Later, inherent from the U-Net framework~\cite{60}, Jiang et al.~\cite{111} propose the W-Net model, which leverages two parallel convolution-based encoders to separately obtain the style and content information. The final image is produced via the deconvolution-based decoder. StarFont~\cite{112} is a multi-domain transfer model designed to create any character within a font style by taking multiple input character images and a target class label as input. However, its restriction lies in the requirement for a fixed subset of characters during training, which means that is unable to generate characters that are not part of the original training dataset. TET-GAN~\cite{114} expands the artistic Chinese characters into the realm of font transfer, which uses both stylization as well as de-stylization subnetworks under the AE framework. Nonetheless, apart from the vast parameters in seven subnetworks (three encoders, two generators, and two discriminators), TET-GAN is not robust to various glyph structures as well. To address these problems, Li et al.~\cite{38} put forward FET-GAN, an innovative end-to-end architecture to implement font transfer among diverse text effects domains. It takes advantage of an encoder to extract the style vector, and then the generator translates the original effect to the target effect indicated via the vector. AGIS-Net~\cite{35} captures shared style features from a reference image set and connects them with the corresponding content features so as to generate the shape and texture images simultaneously through a cooperative training decoder. Moreover, a local texture refinement loss is inserted to equilibrize the samples in few-shot learning. Deep imitator~\cite{113} incorporates a mata-style matrix to cluster and store certain projections of font styles. Besides, it also adds a style-wise embedding within the vector of the Condition Gate Recurrent Unit (CGRU) to predict the probabilistic density of pen tip movement. OCFGNet~\cite{120} adopts a patch-level discriminator to distinguish whether the received character is real or fake. Along with the adversarial loss, it not only uses per-pix loss but also integrates a new loss term SSIM to drive overall optimization. Yet, a classifier with a finite number of categories is obviously not enough to encode arbitrary font content or style. Aoki et al.~\cite{121} apply metric learning to the style feature encoder, encouraging style features of similar glyphs to cluster and those of different styles to diverge in the embedding space. Zhu et al.~\cite{122} utilize the similarity value between the reference characters and the target character to assign a weight for the style features of each reference character. Whereas, for each new style, the network needs to be fine-tuned in a leave-one-out manner. Figure~\ref{Zhuanna} presents some of its generation results.

\begin{figure}[ht]
    \centering
    \includegraphics[width=0.86\linewidth]{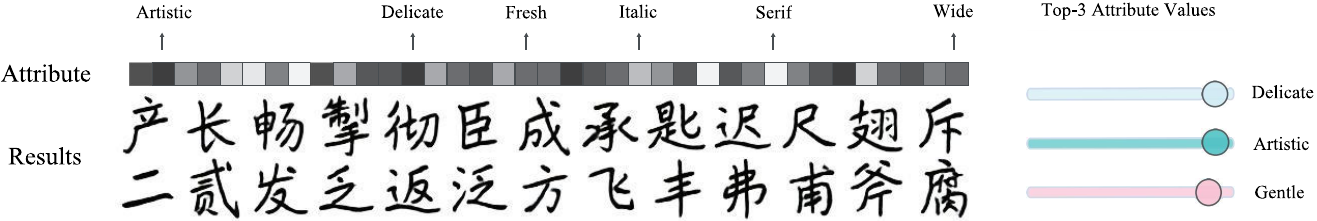}
    \caption{Examples of synthesized Chinese characters under different attribute sets in~\cite{123}.}
    \label{Attribute2Font}
\end{figure}

Subsequently, Wang et al.~\cite{123} raise Attribute2Font that automatically generates visually appealing glyphs based on the user-specified attributes and their respective values, as shown in Figure~\ref{Attribute2Font}. Nevertheless, the assumption that each font corresponds to a set of font attribute values is not completely accurate. In some situations, an exquisite font may exhibit unique characteristics beyond these attributes. As a consequence, once the pre-defined attributes cannot adequately capture a desired font style, the model would certainly fail to portray the font style. Later, motivated by meta-learning, Chen et al.~\cite{116} propose a meta-training strategy based on the Model-Agnostic Meta-Learning (MAML)~\cite{117} along with a task organization method called MLFont for font generation. This strategy facilitates efficient fine-tuning of the generator on new font generation tasks, but the core performance of the model itself is more critical. In response, Chen~\cite{115} presents a new Transformer-based method, which consists of dual encoder modules to learn the character content and font style representation separately, along with a decoder module for generating target characters. Compared with MLFont, Font Transformer benefits from the Transformer's strength in modelling long-range dependencies~\cite{118}. ZiGAN~\cite{119} ingeniously learns the coarse-grained content knowledge from the standard font library. To boost structural understanding, it maps the character features among different styles into Hilbert space and aligns the feature distributions. FTransGAN~\cite{36} introduces two novel modules, context-aware attention network and layer attention network, to synchronously learn the local and global style features. However, it only enables the receiving of a fixed number of style images during the training period due to the architectural design of the model. Founded on the unsupervised method TUNIT~\cite{136}, DG-Font~\cite{124} takes advantage of a feature deformation skip connection (FDSC) module that predicts displacement map pairs and applies deformable convolutions~\cite{167} to the low-level feature maps from the content encoder. Afterwards, DG-Font++~\cite{135} deeper amalgamates local spatial attention into the FDSC module. It can predict similarity scores for each feature position relative to its neighbouring positions, thereby refining the overall output. Sequentially, CF-Font~\cite{130} further extends DG-Font to construct a robust content feature by projecting the content feature into a linear space. It brings in an iterative style-vector refinement strategy to obtain a better feature vector for font-level style representation likewise. In spite of that, it still has difficulty in handling large variations. MF-Net~\cite{126} uses a language complexity-aware skip connection to adaptively coordinate the preservation of structural information from the content. Nevertheless, owing to the specific neural network design, the model can only export the result at a fixed resolution of 64$\times$64.

\begin{figure}[t]%
\centering
\includegraphics[width=0.89\textwidth]{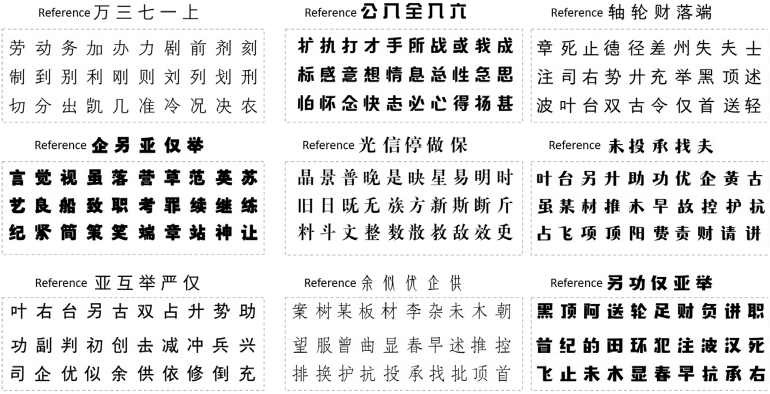}
\caption{Generation results on nine Chinese unseen styles in Hassan et al.~\cite{131}.}\label{Qualitative results on Chinese unseen four styles}
\end{figure}

\begin{figure}[t]
    \centering
\includegraphics[width=0.89\linewidth]{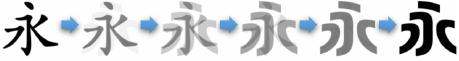}
    \caption{Pixel-level transformation process of font generation in NTF~\cite{129}.}
    \label{NTF}
\end{figure}

\begin{figure}
    \centering
\includegraphics[width=\linewidth]{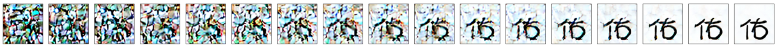}
    \caption{The generation process by the diffusion model~\cite{2}.}
    \label{Diffusion}
\end{figure}

Considering the positive and negative samples' relationships across different styles, DS-Font~\cite{128} leverages a cluster-level contrastive style loss to learn better style representations. Similarly, Hassan et al.~\cite{131} introduce a font style-guided discriminator with three output branches: the first extracts style features to guide the generator, the second produces standard GAN logits, and the third learns a font-style embedding space. Besides, they choose the triplet loss to guarantee that the distance between the anchor and negative is smaller than that between the anchor and the positive during model training. Figure~\ref{Qualitative results on Chinese unseen four styles} demonstrates some of its generation results. However, as the model solely learns the global style representation, it is subject to its capability to represent diverse partial font style shifts. Next, guided by the advances of the Neural Radiance Field (NeRF)~\cite{137} in 3D view synthesis, NTF~\cite{129} treats the Chinese font generation as a continuous transformation process where font pixels are created and dissipated along a transformation path (as shown in Figure~\ref{NTF}), which is then embedded into the NeRF. Meanwhile, a differentiable font rendering procedure is developed to accumulate the intermediate transformations, ultimately producing the target font image. Nevertheless, regardless of whether using global or component-wise disentanglement, an average operation is usually performed on the extracted features, which easily weakens the local information and results in the loss of fine-grained details. LSG-FCST~\cite{37} employs a similarity feature guidance to generate styles within spatial information. It establishes a transfer network that encodes both content and style features from low-level to high-level semantics while using an attention-matching mechanism to calculate the similarity between the style reference image and the source image. Jin et al.~\cite{134} put forward CLF-Net, which captures the inter-image relationships and enables the model to learn essential style features at different character scales through a Multi-scale External Attention Module (MEAM). TransFont~\cite{132} exploits the long-range dependency modelling ability of the ViT~\cite{138}, empirically demonstrating that ViT outperforms CNN in character image generation due to its strength in shape recognition. In addition, it appends a glyph self-attention module to mitigate the quadratic computational and memory overhead. Yang et al.~\cite{2} propose FontDiffuser, a diffusion-based image-to-image one-shot font generation method that frames font generation learning as a noise-to-denoise paradigm, as illutrated in Figure~\ref{Diffusion}. Moreover, they innovatively integrate the Multiscale Content Aggregation (MCA) block, which utilizes both global as well as local content features at different scales. Nevertheless, the model underperforms on the Chinese characters with simple glyphs and structures. Drawn from the common practice of expert font designers, MSD-Font~\cite{127} reformulates the font generation as a multi-stage generative process through incorporating the font transfer process into the diffusion model, with a dual-network approach to better deal with behaviours in different generative stages. Whereas, as it is built upon the diffusion model, MSD-Font has a larger model size and demands more inference time than the earlier GAN-based approaches. Liu et al.~\cite{172} introduce deformable convolutions in the skip connection and design a multi-scale style discriminator to evaluate the style consistency of generated images at different scales.

We present a succinct synopsis of these universal-feature-based generation methods in Table~\ref{universalfeaturerepresentationbasedmethods}. On the plus side, these methods offer strong adaptability, enabling efficient font generation with minimal references and are relatively simple to implement. However, they often encounter difficulties in accurately capturing fine-grained structural details and subtle stylistic nuances, which occasionally result in imprecise or distorted results, especially when dealing with fancy and complex Chinese characters.

\begin{table}[htbp]
\centering
\footnotesize
\caption{A summary of presented universal-feature-based methods.}\label{universalfeaturerepresentationbasedmethods}
\begin{spacing}{1.9}
\vspace{2mm}
\begin{tabular}{lll}
    \toprule
    Literature & Year & Brief description  \\
    \midrule
    \makecell[l]{EMD~\cite{110}}   &  \makecell[l]{2018} &  \makecell[l]{Extract the style and content features from a set of style-reference images and content images separately}    \\
    \makecell[l]{W-Net~\cite{111}}   &  \makecell[l]{2018} &  \makecell[l]{Utilize two parallel convolution-based encoders to capture the style and content information individually}    \\
    \makecell[l]{StarFont~\cite{112}}  & \makecell[l]{2019} & \makecell[l]{Create characters within a font style by taking multiple character images and a target class label as input}   \\
    \makecell[l]{TET-GAN~\cite{114}}  & \makecell[l]{2019} & \makecell[l]{Use both stylization and destylization subnetworks to realize artistic Chinese font transfer based on AE}   \\
    \makecell[l]{AGIS-Net~\cite{35}}  & \makecell[l]{2019} & \makecell[l]{Capture the style features from reference set and connect them with the corresponding content features}   \\
    \makecell[l]{FET-GAN~\cite{38}}  & \makecell[l]{2020} & \makecell[l]{Model each font style's representation as an effect code and utilize it to realize the effect transformation}   \\
    \makecell[l]{Deep imitator~\cite{113}}  & \makecell[l]{2020} & \makecell[l]{Integrate a mata-style matrix into the model to cluster and store the projections of learned writing styles}   \\
    \makecell[l]{OCFGNet~\cite{120}}  & \makecell[l]{2020} & \makecell[l]{Take advantage of a patch-level discriminator to tell apart whether the received character is real or fake}   \\
    \makecell[l]{Aoki et al.~\cite{121}}  & \makecell[l]{2020} & \makecell[l]{Apply the metric learning in the style feature encoder to force the vectors closer if having the same style}   \\
    \makecell[l]{Zhu et al.~\cite{122}}  & \makecell[l]{2020} & \makecell[l]{Employ the similarity value to assign a weight contribution for style features of each reference character} \\
    \makecell[l]{Attribute2Font~\cite{123}}  & \makecell[l]{2020} & \makecell[l]{Generate the visually appealing glyphs based on the user-specified attributes and their respective values}   \\
    \makecell[l]{ZiGAN~\cite{119}}  & \makecell[l]{2021} & \makecell[l]{Learn the coarse-grained content knowledge from standard font library to boost structural understanding}   \\
    \makecell[l]{FTransGAN~\cite{36}}  & \makecell[l]{2021} & \makecell[l]{Bring in context-aware attention and layer attention networks to learn the local and global style features}  \\
    \makecell[l]{MLFont~\cite{116}}  & \makecell[l]{2021} & \makecell[l]{Adopt a meta-training strategy based upon the MAML to facilitate efficient fine-tuning of the generator}   \\
    \makecell[l]{DG-Font~\cite{124}}  & \makecell[l]{2021} & \makecell[l]{Raise a FDSC module to predict displacement map pairs and deformable convolution to content encoder}   \\
    \makecell[l]{DG-Font++~\cite{135}}  & \makecell[l]{2022} & \makecell[l]{Blend the local spatial attention into the FDSC module founded on DG-Font to refine the overall output}   \\
    \makecell[l]{Chen~\cite{115}}  & \makecell[l]{2022} & \makecell[l]{Involve dual encoders to learn the content and style representation along with a decoder to create images}   \\
    \makecell[l]{MF-Net~\cite{126}}  & \makecell[l]{2022} & \makecell[l]{Exploit a language complexity-aware skip connection to adaptively coordinate the structure preservation}   \\
    \makecell[l]{DS-Font~\cite{128}}  & \makecell[l]{2023} & \makecell[l]{Leverage a cluster-level contrastive style loss to promote the feature representation of distinct font types}   \\
    \makecell[l]{NTF~\cite{129}}  & \makecell[l]{2023} & \makecell[l]{Treat font generation as a continuous process where the pixels are processed along a transformation path}   \\
    \makecell[l]{CF-Font~\cite{130}}  & \makecell[l]{2023} & \makecell[l]{Extend DG-Font to construct a robust content feature by projecting the content feature into a linear space}   \\
    \makecell[l]{LSG-FCST~\cite{37}}  & \makecell[l]{2024} & \makecell[l]{Make use of a similarity feature guidance strategy to create diverse font styles within spatial information}   \\
    \makecell[l]{CLF-Net~\cite{134}}  & \makecell[l]{2024} & \makecell[l]{Establish the inter-image relationships and learn useful style features at different scales through MEAM}   \\
    \makecell[l]{Hassan et al.~\cite{131}}  & \makecell[l]{2024} & \makecell[l]{Put forward a font style-guided discriminator with three output branches and a novel triplet loss function}  \\
    \makecell[l]{TransFont~\cite{132}}  & \makecell[l]{2024} & \makecell[l]{Apply the ViT's long-range dependency modelling ability and insert an extra glyph self-attention module}   \\
    \makecell[l]{FontDiffuser~\cite{2}}  & \makecell[l]{2024} & \makecell[l]{Frame font generation as a noise-to-denoise paradigm and use MCA block to learn distinct scale features}   \\
    \makecell[l]{MSD-Font~\cite{127}}  & \makecell[l]{2024} & \makecell[l]{Reformulate font generation as a multi-stage process via incorporating style transfer into diffusion model}   \\
    \makecell[l]{Liu et al.~\cite{172}}  & \makecell[l]{2025} & \makecell[l]{Introduce deformable convolutions in the skip connection and propose a multi-scale style discriminator}   \\
    \bottomrule
\end{tabular}
\end{spacing}
\end{table}

\subsection{Structural-feature-based methods}

\begin{figure}[h]%
\centering
\includegraphics[width=0.75\textwidth]{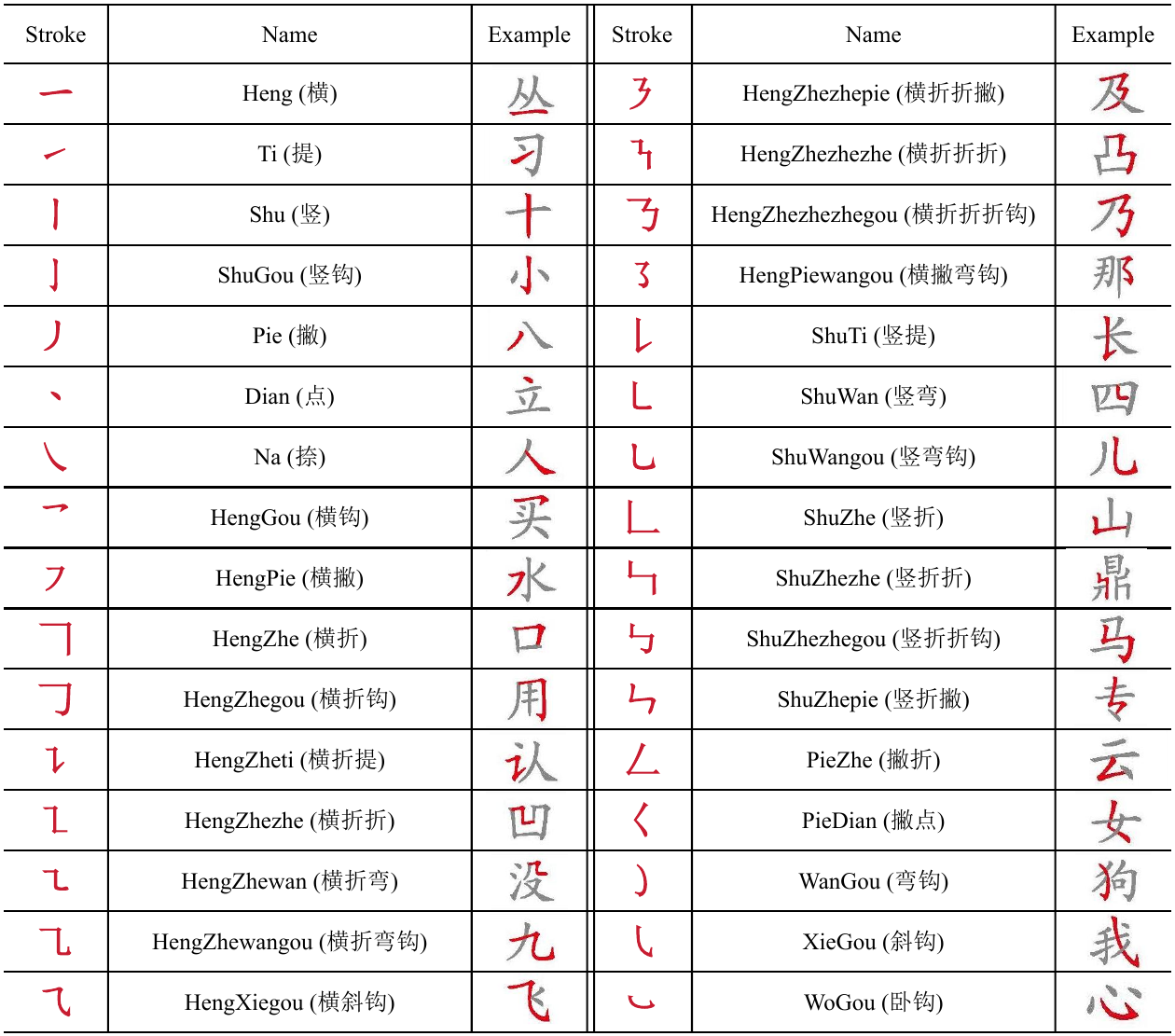}
\caption{Basic strokes of Chinese characters.}
\label{Basic strokes}
\end{figure}

\begin{figure}[!h]
\centering
\includegraphics[width=0.77\textwidth]{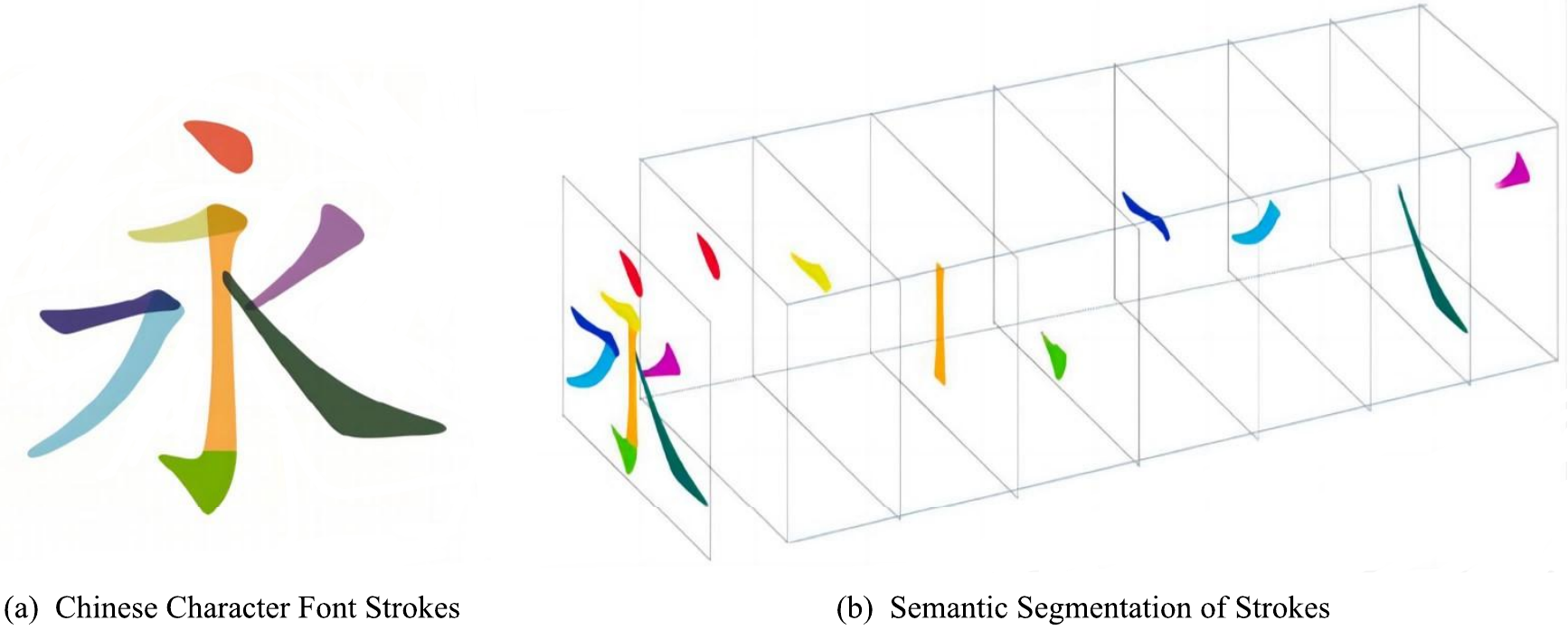}
\caption{A stroke representation of the character sounded as "Yong" in Liu et al.~\cite{77}.}\label{A stroke extraction}
\end{figure}

Witnessing the fine-grained structural variations and local correlations (such as lines, strokes, and components), a quantity of studies adopt structural feature representations to boost performance, which chiefly focus on decomposing Chinese characters according to the structure and obtaining multiple localized style representations. Figure~\ref{Basic strokes} shows 32 basic strokes of Chinese characters, and Figure~\ref{A stroke extraction} exhibits a stroke extraction example of the Chinese character.

SA-VAE~\cite{11} integrates a pre-trained recognition network to provide correct content labels. It encodes the recognized characters into unique codes representing the structural configurations of 12 high-frequency Chinese characters and 101 commonly used radicals. Lian et al.~\cite{139} develop a system called EasyFont, which designs a stroke extraction algorithm to construct the optimal reference data from a pre-trained font skeleton manifold and then establishes correspondences between target and reference characters via a non-rigid point set registration. Whereafter, grounded in the fact that all the Chinese characters share the same radical lexicon, RD-GAN~\cite{140} leverages a new radical extraction module to obtain rough radicals. By separating characters into constituent radicals, the glyph rendering difficulty decreases greatly. Later, to dynamically obtain the style information, Tang et al.~\cite{141} employ large margin softmax loss and add an attention-based adaptive style block before the decoder so as to promote discriminative learning. Whereas, the characters in this model are constrained to the writing trajectory without contour, making it unsuitable for dealing with brush-style characters directly. XMP-Font~\cite{145} raises a self-supervised pre-training strategy alongside a cross-modality transformer-based encoder to model the style feature representations at three levels (stroke-level, component-level, and character-level), thus facilitating the style-content disentanglement. Based on deep reinforcement learning (DRL), FontRL~\cite{142} utilizes a two-stage architecture that first modifies reference stroke skeletons and then predicts stroke bounding boxes to ensure that the generated character skeleton is satisfactory. LF-Font~\cite{143} obtains component-wise style features via employing a component-based style encoder, which facilitates the capture of local details in rich text design. Next, to be less dependent on explicit component annotations, MX-Font~\cite{144} takes advantage of a multi-headed encoder design, where each head can extract different localized features in a weak supervision manner. MX-Font++~\cite{173} further extends MX-Font through incorporating Heterogeneous Aggregation Experts (HAE) and a content-style homogeneity loss to better aggregate the channel and spatial information. CG-GAN~\cite{146} integrates a Component-Aware Module (CAM) to achieve a more refined separation of content and style at a granular level. Diff-Writer~\cite{13} makes use of a character embedding dictionary to store the content information for each character. Following that, Diff-Font~\cite{160} adopts stroke information to support the sampling process of the diffusion model, but it cannot generate unseen characters.

\begin{figure}[h]%
\centering
\includegraphics[width=0.99\textwidth]{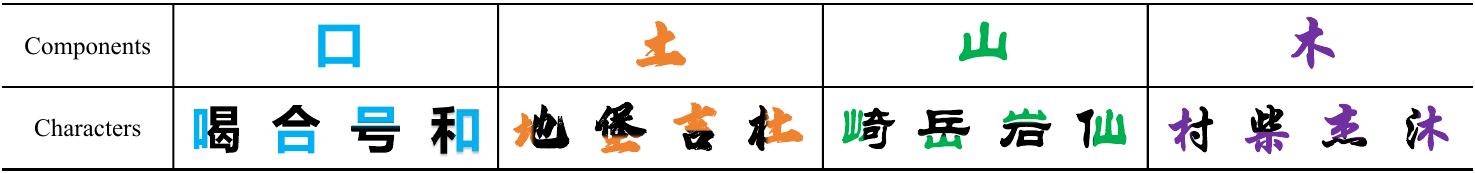}
\caption{Examples of inconsistent contours of the same component in different glyphs across four font styles.}
\label{Examples of inconsistent contours}
\end{figure}

\begin{figure}
    \centering
    \includegraphics[width=0.68\linewidth]{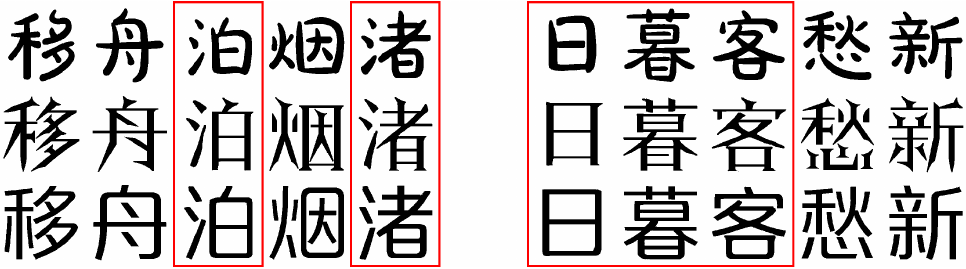}
    \caption{Some generation results of FontTransformer~\cite{150}. Model-synthesized glyphs are marked in the red boxes and the others are designed/witten by human beings.}
    \label{FontTransformer}
\end{figure}

Subsequently, to discover the spatial correspondences between the content images and style images, FsFont~\cite{148} utilizes a cross-attention mechanism to aggregate style features into a fine-grained style representation. However, the model tends to suffer performance degradation if the candidate components have distinct profiles from their counterparts in the target glyph. To solve this issue, considering that the same component also has obvious differences among various characters (as shown in Figure~\ref{Examples of inconsistent contours}), Zhao et al.~\cite{153} leverage informative component pairs to explicitly warp candidate components and align them with the profiles of their counterparts. This tactic ensures more consistent style representations. Yuan et al.~\cite{149} propose SE-GAN, a one-stage model that includes two encoders to catch features from both the source image and its corresponding skeleton image. Besides, a Self-Attentive Refined Attention Module (SAttRAM) is applied in the generator to efficiently fuse the features as well. DeepVecFont~\cite{152} designs a dual-modality learning strategy, which uses image-aspect along with sequence-aspect font features to synthesize Chinese characters. Nonetheless, it heavily leans on image-guided outline refinement during post-processing. Therefore, an updated version, DeepVecFont-v2~\cite{151}, is presented. It employs a Transformer-based generative model along with a relaxation representation of vector outlines so as to synthesize high-quality vector fonts with compact and coordinated outlines. Likewise, for high-resolution (\textit{e.g.}, 1024$\times$1024) image synthesis, Liu et al.~\cite{150} put forward FontTransformer. The core innovation lies in employing a parallel Transformer to prevent the accumulation of prediction errors, while a serial Transformer is adopted to refine and enhance the quality with more compact and cohesive outlines. Figure~\ref{FontTransformer} presents some of its generation results.

\begin{figure}[t]
    \centering
\includegraphics[width=0.51\linewidth]{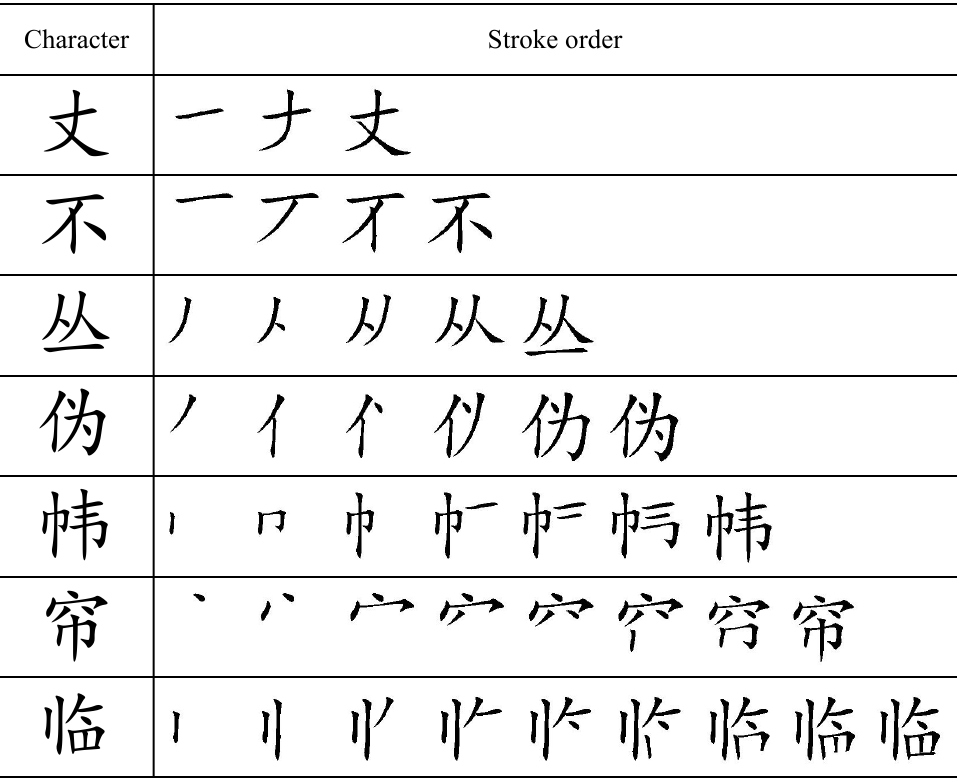}
    \caption{Illustration of stroke writing order.}
    \label{stroke writing order}
\end{figure}

\begin{figure}[t]
    \centering
\includegraphics[width=0.51\linewidth]{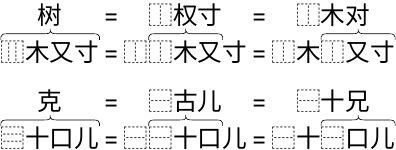}
    \caption{Illustration of equivalent IDSs.}
    \label{IDS}
\end{figure}

Afterwards, Su et al.~\cite{154} bring in a Content-Component Aware Module (CCAM). In addition, a glyph style contrastive enhancement strategy is also fused into the model to strengthen the learning of stable and robust font style representations. EDDNet~\cite{155} decomposes each Chinese character into a series of radical sequences. Simultaneously, it exploits a fine-grained content discriminator to guide the generation procedure, ensuring that the content integrity is properly preserved. SDT~\cite{156} employs a dual-head style encoder and leverages contrastive learning to guide the model in concentrating on writer-wise styles as well as character-wise styles. Pan et al.~\cite{157} assemble styles from character similarity-guided global features and stylized component-level representations. DeepCalliFont~\cite{159} integrates both the information of glyph images and writing trajectories via a dual-modality representation learning, accompanied by a differentiable rasterization loss to ensure consistency between the two modalities. VQ-Font~\cite{165} utilizes a font codebook to encapsulate tokens prior to refining synthesized font images. By virtue of mapping the generated fonts into the discrete space defined by the codebook, VQ-Font can better learn the structural styles of the reference characters. Analogously, based on AE, Liu et al.~\cite{161} quantize the image and sequence modal features using a shared codebook and map them to the same discrete space for realizing the alignment with similar distances and semantics. CLIP-Font~\cite{162} incorporates semantic self-supervision to embed richer semantic information into the model, which could represent the font content more comprehensively. DP-Font~\cite{164} adopts a multi-attribute guidance along with a crucial constraint of stroke writing order (as shown in Figure~\ref{stroke writing order}) to steer the generation process within a diffusion model framework. HFH-Font~\cite{169} employs a diffusion-based generative framework with component-aware conditioning to learn different levels of style information, which is adaptable to varying input image sizes. QT-Font~\cite{163} designs a unique sparse quadtree-based glyph representation to reduce the complexity of the representation space, exhibiting linear complexity and uniqueness. Nevertheless, because the synthesized glyph relies on predictions from multiple network layers, it potentially leads to incorrect results. IF-Font~\cite{168} incorporates Ideographic Description Sequence (IDS) instead of the source glyph to control the semantics of generated glyphs, as described in Figure~\ref{IDS}. EdgeFont~\cite{174} utilizes multi-scale edge information to improve both style and content representations without the need for manual edge labelling. However it is difficult to extend the model to handle calligraphic styles due to the difficulty of capturing stylistically abstracted font features using only edge-based methods.

We provide a pithy recapitulation of these structural-feature-based generation methods in Table~\ref{structurefeaturerepresentationbasedmethods}. In comparison with universal-feature-based methods, these approaches stand out in encapsulating fine-grained local style variations, which enables more precise and flexible font generation, particularly for complex characters and intricate designs. Nevertheless, creating precise annotations and labels for individual components or strokes requires substantial effort and expertise, which would restrict the scalability or automation in practice.

\begin{table}[htbp]
\centering
\footnotesize
\caption{A summary of presented structural-feature-based methods.}\label{structurefeaturerepresentationbasedmethods}
\begin{spacing}{1.8}
\vspace{2mm}
\begin{tabular}{lll}
    \toprule
    Literature & Year & Brief description  \\
    \midrule
    \makecell[l]{SA-VAE~\cite{11}}   &  \makecell[l]{2018} &  \makecell[l]{Bring in a recognition network to offer correct content labels and encode the characters into codes}    \\
    \makecell[l]{EasyFont~\cite{139}}   &  \makecell[l]{2018} &  \makecell[l]{Design a stroke extraction algorithm to construct the reference data from a font skeleton manifold}    \\
    \makecell[l]{RD-GAN~\cite{140}}  & \makecell[l]{2020} & \makecell[l]{Take advantage of a new radical extraction module to separate characters into constituent radicals}   \\
    \makecell[l]{Tang et al.~\cite{141}}  & \makecell[l]{2021} & \makecell[l]{Employ large margin softmax loss and insert an attention-based adaptive block before the decoder}   \\
    \makecell[l]{FontRL~\cite{142}}  & \makecell[l]{2021} & \makecell[l]{Utilize a two-stage model that first modifies the stroke skeletons and then predict bounding boxes}   \\
    \makecell[l]{LF-Font~\cite{143}}  & \makecell[l]{2021} & \makecell[l]{Obtain the component-wise character style features by virtue of a component-based style encoder}   \\
    \makecell[l]{MX-Font~\cite{144}}  & \makecell[l]{2021} & \makecell[l]{Put forward a multi-headed encoder where each head is able to extract different localized features}   \\
    \makecell[l]{DeepVecFont~\cite{152}}  & \makecell[l]{2021} & \makecell[l]{Exploit a dual-modality learning strategy to mix both image-related and sequence-related features}   \\
    \makecell[l]{XMP-Font~\cite{145}}  & \makecell[l]{2022} & \makecell[l]{Use a self-supervised cross-modality pre-training strategy along with a transformer-based encoder}   \\
    \makecell[l]{CG-GAN~\cite{146}}  & \makecell[l]{2022} & \makecell[l]{Integrate CAM to carry out a more refined disentanglement of content and style at a granular level}   \\
    \makecell[l]{FsFont~\cite{148}}  & \makecell[l]{2022} & \makecell[l]{Blend style features from reference glyphs into the fine-grained through a cross-attention module}   \\
    \makecell[l]{SE-GAN~\cite{149}}  & \makecell[l]{2022} & \makecell[l]{Apply two encoders to catch features from the source image and its corresponding skeleton image}  \\
    \makecell[l]{Diff-Font~\cite{160}}  & \makecell[l]{2022} & \makecell[l]{Infuse prior stroke information into the condition diffusion model to support the sampling process}  \\
    \makecell[l]{FontTransformer~\cite{150}}  & \makecell[l]{2023} & \makecell[l]{Put forth a parallel Transformer to strengthen the quality with more compact and cohesive outlines}   \\
    \makecell[l]{DeepVectFont-v2~\cite{151}}  & \makecell[l]{2023} & \makecell[l]{Generate high-quality vector fonts via a Transformer-based model and a representation of outlines}   \\
    \makecell[l]{Su et al.~\cite{154}}  & \makecell[l]{2023} & \makecell[l]{Melt CCAM and a glyph style contrastive enhancement strategy into model to boost style learning}   \\
    \makecell[l]{EDDNet~\cite{155}}  & \makecell[l]{2023} & \makecell[l]{Decompose each character into a series of radical sequences to promote the generation procedure}   \\
    \makecell[l]{SDT~\cite{156}}  & \makecell[l]{2023} & \makecell[l]{Implement a dual-head style encoder and contrastive learning to focus on comprehensive features}   \\
    \makecell[l]{Pan et al~\cite{157}}  & \makecell[l]{2023} & \makecell[l]{Fuse styles from character similarity-guided global features and stylized component-level features}   \\
    \makecell[l]{Diff-Writer~\cite{13}}  & \makecell[l]{2023} & \makecell[l]{Store the content information for each Chinese character through a character embedding dictionary}   \\
    \makecell[l]{VQ-Font~\cite{165}}  & \makecell[l]{2024} & \makecell[l]{Set up a specific font codebook to encapsulate tokens prior to refining the synthesized font images}   \\
    \makecell[l]{Zhao et al.~\cite{153}}  & \makecell[l]{2024} & \makecell[l]{Leverage informative component pairs to warp candidate components and align them with profiles}   \\
    \makecell[l]{DeepCalliFont~\cite{159}}  & \makecell[l]{2024} & \makecell[l]{Combine the image information and writing trajectories via a dual-modality representation learning}   \\
    \makecell[l]{Liu et al.~\cite{161}}  & \makecell[l]{2024} & \makecell[l]{Quantize the image as well as sequence modal features, and map them into the same discrete space}  \\
    \makecell[l]{CLIP-Font~\cite{162}}  & \makecell[l]{2024} & \makecell[l]{Incorporate the semantic self-supervision to provide richer semantic information within the model}   \\
    \makecell[l]{DP-Font~\cite{164}}  & \makecell[l]{2024} & \makecell[l]{Adopt a multi-attribute guidance along with a crucial constraint of stroke order to steer the process}   \\
    \makecell[l]{HFT-Font~\cite{169}}  & \makecell[l]{2024} & \makecell[l]{Apply a diffusion-based model with component-aware conditioning to learn styles at diverse levels}   \\
    \makecell[l]{QT-Font~\cite{163}}  & \makecell[l]{2024} & \makecell[l]{Propose a unique sparse quadtree-based character representation to reduce the model's complexity}   \\
    \makecell[l]{IF-Font~\cite{168}}  & \makecell[l]{2024} & \makecell[l]{Incorporate Ideographic Description Sequence (IDS) to control the semantics of generated glyphs}   \\
    \makecell[l]{MX-Font++~\cite{173}}  & \makecell[l]{2025} & \makecell[l]{Extend MX-Font via incorporating Heterogeneous Aggregation Experts and a homogeneity loss}   \\
    \makecell[l]{EdgeFont~\cite{174}}  & \makecell[l]{2025} & \makecell[l]{Utilize multi-scale edge information to improve both the style and content feature representations}   \\
    \bottomrule
\end{tabular}
\end{spacing}
\end{table}

In Table~\ref{summary of few shot methods}, we summarize the two sorts of few-shot Chinese font generation methods discussed above. While these methods can produce high-quality font images, it is a tough task to achieve a complete disentanglement between the content and style features. Apart from this, the font selected for content-feature encoding plays a critical role in shaping the final results. For instance, fonts like Song and Kai are frequently chosen as the content font. Although such choices are proven to be effective in many cases, the generated images sometimes exhibit artifacts, including incomplete lines or unwanted radicals. 

\begin{table}[htbp]
\centering
\footnotesize
\caption{Classification of few-shot Chinese font generation approaches.}\label{summary of few shot methods}
\begin{spacing}{1.9}
\vspace{2mm}
\begin{tabular}{llll}
    \toprule
    Type of method & Simplistic depiction & Strengths & Limitations   \\
    \midrule
    \makecell[l]{Universal-feature-based}  &  \makecell[l]{Synthesize glyphs by merging the style \\ features and the content features directly} &  \makecell[l]{Offer good adaptability and \\relatively simple to deploy}  & \makecell[l]{Fail to capture fine-grained structural \\details and subtle stylistic nuances}   \\ 
    \midrule[0.25pt]
    \makecell[l]{Structural-feature-based}   &  \makecell[l]{Decompose Chinese characters to obtain \\ multiple localized style representation} &  \makecell[l]{Excel in extracting local style \\features and variations}  & \makecell[l]{Labelling precise annotations requires \\ substantial effort and expertise}  \\
    \bottomrule
\end{tabular}
\end{spacing}
\end{table}

\section{Discussion}\label{Discussion}
\subsection{Challenges of deep learning for Chinese font generation}
\subsubsection{Intricate glyph structure and numerous characters}
As we have pointed out above, unlike other alphabetic systems, Chinese characters are composed of a diverse array of strokes, radicals and spatial arrangements, which makes it hard to encapsulate and replicate the finer structural details during the generation process. Considering the vast number of Chinese characters, the difficulty intensifies when aiming to create high-quality fonts that preserve stylistic consistency across thousands of unique characters. Apart from this, some Chinese fonts like cursive script, are highly personalized and random, even the same font can exhibit significant style differences as written by different people~\cite{105}. Therefore, precisely extracting and learning these elaborate and personalized styles, especially in the case of finite available data, remains an intractable problem in promoting the generation techniques. While attention mechanisms~\cite{103,149,156,159} have shown promise in enhancing style transfer, an excessive focus on complex cursive styles may compromise the fidelity of standard script generation. A potential solution lies in introducing dedicated modules for cursive style modelling, enabling targeted learning and transformation without degrading the performance on other styles.

\subsubsection{Limited availability of high-quality datasets}
Although we have outlined six publicly available datasets in subsection~\ref{datasetsection}, some of them are restricted in scope or diversity, failing to cover the extensive range of Chinese characters and styles. Besides, in most studies, the reliance on custom or private datasets constrains the multiformity of data because researchers usually have to curate their own collections. Furthermore, since digital fonts fall under the category of creative intellectual property, the copyright restrictions further hinder the realization of open sharing because there are concerns about unauthorized copying, alteration and misuse. These limitations not only hinder the data diversity but also constrain models’ ability to generalize to unseen characters and styles. Therefore, developing comprehensive datasets that encompass a wide variety of characters and styles is an important factor in enhancing the models' effectiveness. 

\subsubsection{Quantification of evaluation metrics}
The evaluation metrics constitute the core part of the Chinese font generation task. As of now, how to accurately assess the effects of output images is also a problem worthy of in-depth exploration. While some widely-used quantitative metrics (including MSE, SSIM and PNSR) can offer basic assessments between the generated glyphs and ground truth, they often fail to capture the nuanced visual characteristics that define the essence of Chinese fonts. These shortcomings are particularly evident in calligraphic texts, where subtle features, such as ink penetration vitality, stroke coherence and dynamic brushwork, play a decisive role in aesthetic perception. Moreover, the lack of benchmark datasets and unified testing mechanisms, together with the inherently subjective nature of font aesthetics and the diversity of individual preferences, leads to substantial inconsistencies in the assessment results~\cite{74}. Bridging this gap requires more robust and culturally informed evaluation frameworks that can reconcile computational metrics with human perceptual judgments.

\subsubsection{High computational complexity and resource requirements}
Nowadays, deep learning techniques for Chinese font generation often entail substantial computational overhead and prolonged inference times, particularly when handling high-resolution character images or complex glyph structures. For instance, models such as FontDiffuser~\cite{2} require approximately 6 seconds to generate a single Chinese character on an RTX 3090 GPU. Such demands present significant challenges for researchers and practitioners with constrained computational resources and hinder the deployment in resource-limited environments. Moreover, the escalating complexity of state-of-the-art architectures (including large-scale transformers, diffusion processes or multi-stage pipelines) further exacerbates the trade-off between quality and efficiency. Therefore, advancing algorithmic efficiency, reducing model size and accelerating inference time without compromising generation quality remain critical directions for future research in this field.

\subsection{Future research directions}
\subsubsection{Practice of network compression strategies}
With the rapid growth in network parameters, both the training and inference phases of generation models are becoming increasingly time-consuming and resource-intensive. Hence, applying network compression strategies offers a promising pathway to improve efficiency. Techniques such as model quantization can reduce the bit-width representation of parameters (\textit{e.g.}, weights and activations), thereby lowering memory consumption and computational cost. Similarly, knowledge distillation enables the transfer of essential style-content knowledge from a large model to a lightweight student network, retaining key generative capabilities while substantially reducing complexity. Furthermore, future explorations could also integrate compression with font-specific structural priors, allowing the compact model to preserve fine-grained structural fidelity even under aggressive size reduction.

\subsubsection{Multimodal learning of Chinese font generation}
Multimodal learning frameworks that integrate images, text and stroke-based information hold great promise for tackling a wider range of generation tasks, including the largely unexplored direction on creating fonts from textual descriptions. Advancing this line of work would require effective cross-modal alignment strategies, such as incorporating alignment modules or intermediate-layer interactions to ensure mutual guidance between image and sequence branches. Besides, Transformer-based architectures with adaptive alignment and cross-modal attention mechanisms may also enhance the model’s capacity to capture writing styles and fine-grained stroke details. In addition, some complementary approaches, such as self-supervised pre-training on unlabelled stroke sequence data (\textit{e.g.}, contrastive learning or masked sequence reconstruction) and targeted data augmentation could improve both robustness and generalization, ultimately producing characters that are more faithful to practical applications.

\subsubsection{Cross-lingual font generation}
Cross-lingual font generation is a compelling research avenue, which enables models to create Chinese fonts based upon the inputs from other languages or scripts. Although there have already been some endeavours in this area~\cite{36,126,134}, they often suffer from issues such as distorted structures or stroke errors, making it difficult to balance content fidelity and stylistic completeness. Future research could leverage prior knowledge of character components, stroke patterns or morphemic structures in the target language to better preserve content information during transfer. Besides, incorporating multi-scale or adaptive attention mechanisms may further enhance style learning by capturing global stylistic patterns while refining fine-grained details. Moreover, large-scale unlabelled cross-lingual font data also presents an opportunity for self-supervised pretraining, enabling the extraction of universal style elements and component features that can be fine-tuned for downstream cross-lingual tasks. Such advances would not only improve model adaptability but also foster inclusivity and cultural relevance in digital typography.

\section{Conclusion}\label{Conclusion}
In this paper, we review the recent advancements in Chinese font generation since the deep learning era. To ensure a comprehensive survey, we depict the methodology utilized for selecting and analyzing the relevant literature. Built upon the collected studies, we introduce the related fundamentals for Chinese font generation, involving classical deep learning architectures, font representation formats, public datasets and commonly-used evaluation metrics. We then separate existing methods into two primary categories: many-shot generation and few-shot generation methods. For each category, we present the classical approaches and analyze their merits and drawbacks, respectively. Based on our investigation, we ultimately conclude with a summary of the challenges faced by this task, and discuss several promising research directions, with the prospect to provide beneficial and fresh insights for studying Chinese font generation in the coming future.

\section*{CRediT authorship contribution statement}
\textbf{Weiran Chen:} Writing - original draft, Writing - review \& editing, Methodology, Visualization, Conceptualization, Data curation. \textbf{Guiqian Zhu:} Writing - review \& editing, Investigation, Resources, Formal analysis. \textbf{Ying Li:} Writing - review \& editing, Conceptualization, Supervision, Validation. \textbf{Yi Ji:} Supervision, Validation. \textbf{Chunping Liu:} Supervision, Project administration, Funding acquisition.

\section*{Declaration of competing interest}
The authors declare that they have no known competing financial interests or personal relationships that could have appeared to influence the work reported in this paper.

\section*{Data availability}
No data was used for the research described in the article.

\section*{Acknowledgments}
This work was partially supported by Postgraduate Research \& Practice Innovation Program of Jiangsu Province KYCX24\_3320, Project Funded by the Priority Academic Program Development of Jiangsu Higher Education Institutions, and National Natural Science Foundation of China (NSFC Grant No. 62376041).









\bibliographystyle{elsarticle-num}  
\bibliography{References}

\begin{thebibliography}{100}
\expandafter\ifx\csname url\endcsname\relax
  \def\url#1{\texttt{#1}}\fi
\expandafter\ifx\csname urlprefix\endcsname\relax\def\urlprefix{URL }\fi
\expandafter\ifx\csname href\endcsname\relax
  \def\href#1#2{#2} \def\path#1{#1}\fi

\bibitem{1}
J.~Fei, The evolution of the form and carrier of {C}hinese characters and the historical and cultural inheritance, Sinogram Culture. 335~(11) (2023) 93--95, \href{https://doi.org/10.14014/j.cnki.cn11-2597/g2.2023.11.018}{https://doi.org/10.14014/j.cnki.cn11-2597/g2.2023.11.018}.

\bibitem{2}
Z.~Yang, D.~Peng, Y.~Kong, Y.~Zhang, C.~Yao, L.~Jin, {F}ont{D}iffuser: {O}ne-{S}hot {F}ont {G}eneration via {D}enoising {D}iffusion with {M}ulti-{S}cale {C}ontent {A}ggregation and {S}tyle {C}ontrastive {L}earning, in: Proceedings of the Thirty-Eighth {AAAI} Conference on Artificial Intelligence, {AAAI}, Vancouver, Canada, February 20-27, 2024, pp. 6603--6611, \href{https://doi.org/10.1609/aaai.v38i7.28482}{https://doi.org/10.1609/aaai.v38i7.28482}.

\bibitem{3}
J.~Zhang, G.~Mao, H.~Lin, J.~Yu, C.~Zhou, {O}utline {F}ont {G}enerating from {I}mages of {A}ncient {C}hinese {C}alligraphy, Trans. Edutainment. 5 (2011) 122--131, \href{https://doi.org/10.1007/978-3-642-18452-9\_10}{https://doi.org/10.1007/978-3-642-18452-9\_10}.

\bibitem{4}
B.~Zhou, W.~Wang, Z.~Chen, Easy generation of personal {C}hinese handwritten fonts, in: Proceedings of the 2011 {IEEE} International Conference on Multimedia and Expo, {ICME}, Barcelona, Catalonia, Spain, July 11-15, 2011, pp. 1--6, \href{https://doi.org/10.1109/ICME.2011.6011892}{https://doi.org/10.1109/ICME.2011.6011892}.

\bibitem{5}
A.~Zong, Y.~Zhu, {S}troke{B}ank: {A}utomating {P}ersonalized {C}hinese {H}andwriting {G}eneration, in: Proceedings of the Twenty-Eighth AAAI Conference on Artificial Intelligence, {AAAI}, Qu{\'{e}}bec City, Qu{\'{e}}bec, Canada, July 27-31, 2014, pp. 3024--3029, \href{https://doi.org/10.1609/aaai.v28i2.19029}{https://doi.org/10.1609/aaai.v28i2.19029}.

\bibitem{6}
Z.~Lian, J.~Xiao, Automatic shape morphing for {C}hinese characters, in: Proceedings of SIGGRAPH Asia 2012 Technical Briefs, Singapore, November 28 - December 1, 2012, pp. 1--4, \href{https://doi.org/10.1145/2407746.2407748}{https://doi.org/10.1145/2407746.2407748}.

\bibitem{7}
R.~Cheng, X.~Zhao, H.~Zhou, H.~Ye, Review of {C}hinese font style transfer research based on deep learning, Journal of Zhejiang University (Engineering Science). 56~(3) (2022) 510--519, \href{https://doi.org/10.3785/j.issn.1008-973X.2022.03.010}{https://doi.org/10.3785/j.issn.1008-973X.2022.03.010}.

\bibitem{8}
T.~T. Khoei, H.~O. Slimane, N.~Kaabouch, Deep learning: systematic review, models, challenges, and research directions, Neural Comput. Appl. 35~(31) (2023) 23103–23124, \href{https://doi.org/10.1007/s00521-023-08957-4}{https://doi.org/10.1007/s00521-023-08957-4}.

\bibitem{9}
L.~Alzubaid, J.~Zhang, A.~J. Humaidi, et~al., Review of deep learning: concepts, {CNN} architectures, challenges, applications, future directions, J. Big Data. 8~(1) (2021) 1–74, \href{https://doi.org/10.1186/s40537-021-00444-8}{https://doi.org/10.1186/s40537-021-00444-8}.

\bibitem{10}
W.~Chen, C.~Liu, Y.~Ji, Chinese {C}haracter {S}tyle {T}ransfer {M}odel {B}ased on {C}onvolutional {N}eural {N}etwork, in: Proceedings of the 31st International Conference on Artificial Neural Networks, {ICANN}, Bristol, UK, September 6-9, 2022, pp. 558--569, \href{https://doi.org/10.1007/978-3-031-15937-4\_47}{https://doi.org/10.1007/978-3-031-15937-4\_47}.

\bibitem{11}
D.~Sun, T.~Ren, C.~Li, H.~Su, J.~Zhu, Learning to {W}rite {S}tylized {C}hinese {C}haracters by {R}eading a {H}andful of {E}xamples, in: Proceedings of the Twenty-Seventh International Joint Conference on Artificial Intelligence, {IJCAI}, Stockholm, Sweden, July 13-19, 2018, pp. 920--927, \href{https://doi.org/10.24963/ijcai.2018/128}{https://doi.org/10.24963/ijcai.2018/128}.

\bibitem{12}
J.~Liu, C.~Gu, J.~Wang, G.~Youn, J.-U. Kim, Multi-scale multi-class conditional generative adversarial network for handwritten character generation, J. Supercomput. 75~(4) (2019) 1922–1940, \href{https://doi.org/10.1007/s11227-017-2218-0}{https://doi.org/10.1007/s11227-017-2218-0}.

\bibitem{13}
M.~Ren, Y.~Zhang, Q.~Wang, F.~Yin, C.~Liu, Diff-{W}riter: {A} {D}iffusion {M}odel-{B}ased {S}tylized {O}nline {H}andwritten {C}hinese {C}haracter {G}enerator, in: Proceedings of the 30th International Conference on Neural Information Processing, {ICONIP}, Changsha, China, November 20-23, 2023, pp. 86--100, \href{https://doi.org/10.1007/978-981-99-8141-0\_7}{https://doi.org/10.1007/978-981-99-8141-0\_7}.

\bibitem{14}
X.~Wang, C.~Li, Z.~Sun, L.~Hui, Review of {GAN}-{B}ased {R}esearch on {C}hinese {C}haracter {F}ont {G}eneration, Chinese Journal of Electronics. 33~(3) (2024) 584--600, \href{https://doi.org/10.23919/cje.2022.00.402}{https://doi.org/10.23919/cje.2022.00.402}.

\bibitem{15}
C.~Wang, G.~Wu, Y.~Yao, Y.~Ren, et~al., Review of {C}hinese characters generation and font transfer based on deep learning, Journal of image and Graphics. 27~(12) (2022) 3415--3428.

\bibitem{16}
Z.~Huang, Q.~Chen, W.~Luo, {C}hinese {C}haracter {G}eneration {M}ethod {B}ased on {D}eep {L}earning, Computer Engineering and Applications. 57~(17) (2021) 29--36, \href{https://doi.org/10.3778/j.issn.1002-8331.2103-0297}{https://doi.org/10.3778/j.issn.1002-8331.2103-0297}.

\bibitem{17}
Y.~Ma, Y.~Dong, et~al., A {S}urvey of {C}hinese {C}haracter {S}tyle {T}ransfer, in: Proceedings of the 14th Conference on Image and Graphics Technologies and Applications, {IGTA}, Beijing, China, April 19-20, 2019, pp. 392--404, \href{https://doi.org/10.1007/978-981-13-9917-6\_38}{https://doi.org/10.1007/978-981-13-9917-6\_38}.

\bibitem{170}
L.~Wang, Y.~Liu, M.~Y. Sharum, R.~Y. et~al., {D}eep learning for {C}hinese font generation: {A} survey, Expert Syst. Appl. 276 (2025) 1--23, \href{https://doi.org/10.1016/j.eswa.2025.127105}{https://doi.org/10.1016/j.eswa.2025.127105}.

\bibitem{171}
Z.~Ren, Y.~Pan, J.~Chen, L.~Zhao, et~al., {A} {S}urvey on {D}eep {L}earning-{B}ased {C}hinese {F}ont {S}tyle {T}ransfer, IEEE Transactions on Artificial Intelligence 1 (2025) 1--16, \href{https://doi.org/10.1109/TAI.2025.3574300}{https://doi.org/10.1109/TAI.2025.3574300}.

\bibitem{18}
M.~J. Page, J.~E. McKenzie, et~al., The {PRISMA} 2020 statement: {A}n updated guideline for reporting systematic reviews, International Journal of Surgery. 88 (2021) 1--9, \href{https://doi.org/10.1016/j.ijsu.2021.105906}{https://doi.org/10.1016/j.ijsu.2021.105906}.

\bibitem{19}
F.~Zhou, L.~Jin, J.~Dong, {R}eview of {C}onvolutional {N}eural {N}etwork, Chinese Journal of Computers. 40~(6) (2017) 1229--1251.

\bibitem{22}
J.~Lai, X.~Wang, Q.~Xiang, Y.~Song, W.~Quan, Review on autoencoder and its application, Journal on Communications. 42~(9) (2021) 218--230, \href{https://doi.org/10.11959/j.issn.1000-436x.2021160}{https://doi.org/10.11959/j.issn.1000-436x.2021160}.

\bibitem{20}
D.~Rumelhart, G.~Hinton, R.~Williams, Learning representations by back-propagating errors, Nature. 323 (1986) 533--536, \href{https://doi.org/10.1038/323533a0}{https://doi.org/10.1038/323533a0}.

\bibitem{21}
H.~Bourlard, Y.~Kamp, Auto-association by multilayer perceptrons and singular value decomposition, Biol. Cybern. 59 (1988) 291--294, \href{https://doi.org/10.1007/BF00332918}{https://doi.org/10.1007/BF00332918}.

\bibitem{23}
D.~P. Kingma, M.~Welling, {A}uto-{E}ncoding {V}ariational {B}ayes, in: Proceedings of 2nd International Conference on Learning Representations, {ICLR}, Banff, AB, Canada, April 14-16, 2014, pp. 1--14.

\bibitem{24}
F.~Xiao, The style-transfer of Chinese character based on Deep Learning, Jinan University, Guangdong, 2018.

\bibitem{25}
I.~J. Goodfellow, J.~Pouget{-}Abadie, M.~Mirza, et~al., {G}enerative {A}dversarial {N}ets, in: Proceedings of the Advances in Neural Information Processing Systems 27: Annual Conference on Neural Information Processing Systems, {NIPS}, Montreal, Quebec, Canada, December 8-13, 2014, pp. 2672--2680.

\bibitem{26}
M.~Mirza, S.~Osindero, Conditional {G}enerative {A}dversarial {N}ets, preprint at \href{http://arxiv.org/abs/1411.1784}{http://arxiv.org/abs/1411.1784} (2014).

\bibitem{27}
J.~Zhu, T.~Park, et~al., Unpaired {I}mage-to-{I}mage {T}ranslation {U}sing {C}ycle-{C}onsistent {A}dversarial {N}etworks, in: Proceedings of {IEEE} International Conference on Computer Vision, {ICCV}, Venice, Italy, October 22-29, 2017, pp. 2242--2251, \href{https://doi.org/10.1109/ICCV.2017.244}{https://doi.org/10.1109/ICCV.2017.244}.

\bibitem{28}
M.~Arjovsky, S.~Chintala, L.~Bottou, Wasserstein {GAN}, preprint at \href{http://arxiv.org/abs/1701.07875}{http://arxiv.org/abs/1701.07875} (2017).

\bibitem{49}
I.~Gulrajani, F.~Ahmed, M.~Arjovsky, V.~Dumoulin, A.~C. Courville, Improved {T}raining of {W}asserstein {GAN}s, in: Proceedings of Advances in Neural Information Processing Systems 30: Annual Conference on Neural Information Processing Systems, {NIPS}, Long Beach, CA, {USA}, December 4-9, 2017, pp. 5767--5777.

\bibitem{93}
Y.~Choi, M.~Choi, M.~Kim, J.~Ha, S.~Kim, J.~Choo, Star{GAN}: {U}nified {G}enerative {A}dversarial {N}etworks for {M}ulti-{D}omain {I}mage-to-{I}mage {T}ranslation, in: Proceedings of {IEEE} Conference on Computer Vision and Pattern Recognition, {CVPR}, Salt Lake City, UT, USA, June 18-22, 2018, pp. 8789--8797, \href{https://doi.org/10.1109/CVPR.2018.00916}{https://doi.org/10.1109/CVPR.2018.00916}.

\bibitem{30}
J.~Zeng, Q.~Chen, Y.~Liu, et~al., Stroke{GAN}: {R}educing {M}ode {C}ollapse in {C}hinese {F}ont {G}eneration via {S}troke {E}ncoding, in: Proceedings of Thirty-Fifth {AAAI} Conference on Artificial Intelligence, {AAAI}, Virtual Event, February 2-9, 2021, pp. 3270--3277, \href{https://doi.org/10.1609/aaai.v35i4.16438}{https://doi.org/10.1609/aaai.v35i4.16438}.

\bibitem{31}
C.~Liang, Research and Application of Font Style Migration Algorithm Based on Deep Learning, Xijing University, Shanxi, 2021.

\bibitem{97}
J.~Zeng, Q.~Chen, M.~Wang, Diversity {R}egularized {S}tar{GAN} for {M}ulti-style {F}onts {G}eneration of {C}hinese {C}haracters, J. Phys.: Conf. Ser. 1880 (2021) 1--10, \href{https://doi.org/10.1088/1742-6596/1880/1/012017}{https://doi.org/10.1109/T10.1088/1742-6596/1880/1/012017}.

\bibitem{118}
A.~Vaswani, N.~Shazeer, N.~Parmar, et~al., Attention is {A}ll you {N}eed, in: Proceedings of Advances in Neural Information Processing Systems 30: Annual Conference on Neural Information Processing Systems, {NIPS}, Long Beach, CA, {USA}, December 4-9,, 2017, pp. 5998--6008.

\bibitem{138}
A.~Dosovitskiy, L.~Beyer, A.~Kolesnikov, et~al., {A}n {I}mage is {W}orth 16x16 {W}ords: {T}ransformers for {I}mage {R}ecognition at {S}cale, in: Proceedings of the 9th International Conference on Learning Representations, {ICLR}, Virtual Event, Austria, May 3-7, 2021, pp. 1--22.

\bibitem{132}
X.~Chen, L.~Wu, Y.~Su, L.~Meng, X.~Meng, Font transformer for few-shot font generation, Comput. Vis. Image Underst. 245 (2024) 1--10, \href{https://doi.org/10.1016/j.cviu.2024.104043}{https://doi.org/10.1016/j.cviu.2024.104043}.

\bibitem{161}
Y.~Liu, F.~Khalid, M.~R. Mustaffa, A.~bin Azman, Dual-modality learning and transformer-based approach for high-quality vector font generation, Expert Syst. Appl. 240 (2024) 1--19, \href{https://doi.org/10.1016/j.eswa.2023.122405}{https://doi.org/10.1016/j.eswa.2023.122405}.

\bibitem{32}
J.~Sohl{-}Dickstein, E.~A. Weiss, N.~Maheswaranathan, S.~Ganguli, Deep {U}nsupervised {L}earning using {N}onequilibrium {T}hermodynamics, in: Proceedings of the 32nd International Conference on Machine Learning, {ICML}, Lille, France, July 6-11, 2015, pp. 2256--2265.

\bibitem{33}
J.~Ho, A.~Jain, P.~Abbeel, {D}enoising {D}iffusion {P}robabilistic {M}odels, in: Proceedings of Advances in Neural Information Processing Systems 33: Annual Conference on Neural Information Processing Systems, {NeurIPS}, December 6-12, virtual, 2020, pp. 1--25.

\bibitem{157}
W.~Pan, A.~Zhu, X.~Zhou, et~al., Few shot font generation via transferring similarity guided global style and quantization local style, in: Proceedings of {IEEE/CVF} International Conference on Computer Vision, {ICCV}, Paris, France, October 1-6, 2023, pp. 19449--19459, \href{https://doi.org/10.1109/ICCV51070.2023.01787}{https://doi.org/10.1109/ICCV51070.2023.01787}.

\bibitem{165}
M.~Yao, Y.~Zhang, X.~Lin, et~al., {VQ-F}ont: {F}ew-{S}hot {F}ont {G}eneration {w}ith {S}tructure-{A}ware {E}nhancement and {Q}uantization, in: Proceedings of Thirty-Eighth {AAAI} Conference on Artificial Intelligence, {AAAI}, Vancouver, Canada, February 20-27, 2024, pp. 16407--16415, \href{https://doi.org/10.1609/aaai.v38i15.29577}{https://doi.org/10.1609/aaai.v38i15.29577}.

\bibitem{160}
H.~He, X.~Chen, C.~Wang, J.~Liu, B.~Du, et~al., Diff-{F}ont: {D}iffusion {M}odel for {R}obust {O}ne-{S}hot {F}ont {G}eneration, preprint at \href{https://arxiv.org/abs/2212.05895}{https://arxiv.org/abs/2212.05895} (2022).

\bibitem{168}
W.~G. Xinpeng~Chen, Xiao~Ke, {IF-F}ont: {I}deographic {D}escription {S}equence-{F}ollowing {F}ont {G}eneration, in: Proceedings of the Advances in Neural Information Processing Systems 38: Annual Conference on Neural Information Processing Systems, {NIPS}, Vancouver, Canada, December 10-15, 2024, pp. 1--23.

\bibitem{151}
Y.~Wang, Y.~Wang, L.~Yu, Y.~Zhu, Z.~Lian, Deep{V}ec{F}ont-v2: {E}xploiting {T}ransformers to {S}ynthesize {V}ector {F}onts with {H}igher {Q}uality, in: Proceedings of {IEEE/CVF} Conference on Computer Vision and Pattern Recognition, {CVPR}, Vancouver, BC, Canada, June 17-24, 2023, pp. 18320--18328, \href{https://doi.org/10.1109/CVPR52729.2023.01757}{https://doi.org/10.1109/CVPR52729.2023.01757}.

\bibitem{152}
Y.~Wang, Z.~Lian, Deep{V}ec{F}ont: {S}ynthesizing {H}igh-quality {V}ector {F}onts via {D}ual-modality {L}earning, {ACM} Trans. Graph. 40~(6) (2021) 1--15, \href{https://doi.org/10.1145/3478513.3480488}{https://doi.org/10.1145/3478513.3480488}.

\bibitem{163}
Y.~Liu, Z.~Lian, Q{T-F}ont: {H}igh-efficiency {F}ont {S}ynthesis via {Q}uadtree-based {D}iffusion {M}odels, in: Proceedings of {ACM} {SIGGRAPH} 2024 Conference Papers, Denver, CO, USA, July 27 - August 1, 2024, pp. 1--11, \href{https://doi.org/10.1145/3641519.3657451}{https://doi.org/10.1145/3641519.3657451}.

\bibitem{169}
Z.~L. Hua~Li, {HFH-F}ont: {F}ew-shot {C}hinese {F}ont {S}ynthesis with {H}igher {Q}uality, {F}aster {S}peed, and {H}igher {R}esolution, in: Proceedings of {SIGGRAPH} Asia Technical Briefs, Tokyo, Japan, December 3-6, 2024, pp. 1--16.

\bibitem{34}
C.~Liu, F.~Yin, et~al., {CASIA} {O}nline and {O}ffline {C}hinese {H}andwriting {D}atabases, in: Proceedings of International Conference on Document Analysis and Recognition, {ICDAR}, Beijing, China, September 18-21, 2011, pp. 37--41, \href{https://doi.org/10.1109/ICDAR.2011.17}{https://doi.org/10.1109/ICDAR.2011.17}.

\bibitem{35}
Y.~Gao, Y.~Guo, Z.~Lian, Y.~Tang, J.~Xiao, Artistic glyph image synthesis via one-stage few-shot learning, ACM Trans. Graph. 38~(6) (2019) 1--12, \href{https://doi.org/10.1145/3355089.3356574}{https://doi.org/10.1145/3355089.3356574}.

\bibitem{38}
W.~Li, Y.~He, Y.~Qi, Z.~Li, Y.~Tang, {FET-GAN}: {F}ont and {E}ffect {T}ransfer via {K}-shot {A}daptive {I}nstance {N}ormalization, in: Proceedings of the Thirty-Fourth {AAAI} Conference on Artificial Intelligence, {AAAI}, New York, NY, USA, February 7-12, 2020, pp. 1717--1724, \href{https://doi.org/10.1609/aaai.v34i02.5535}{https://doi.org/10.1609/aaai.v34i02.5535}.

\bibitem{36}
C.~Li, Y.~T.~M. Lu, et~al., Few-shot {F}ont {S}tyle {T}ransfer between {D}ifferent {L}anguages, in: Proceedings of {IEEE} Winter Conference on Applications of Computer Vision, {WACV}, Waikoloa, HI, USA, January 3-8, 2021, pp. 433--442, \href{https://doi.org/10.1109/WACV48630.2021.00048}{https://doi.org/10.1109/WACV48630.2021.00048}.

\bibitem{39}
S.~Yang, W.~Wang, J.~Liu, {TE141K}: {A}rtistic {T}ext {B}enchmark for {T}ext {E}ffect {T}ransfer, {IEEE} Trans. Pattern Anal. Mach. Intell. 43~(10) (2021) 3709--3723, \href{https://doi.org/10.1109/TPAMI.2020.2983697}{https://doi.org/10.1109/TPAMI.2020.2983697}.

\bibitem{37}
Y.~Li, G.~Lin, M.~He, D.~Yuan, K.~Liao, Layer similarity guiding few-shot {C}hinese style transfer, Vis Comput. 40~(4) (2024) 2265--2278, \href{https://doi.org/10.1007/s00371-023-02915-w}{https://doi.org/10.1007/s00371-023-02915-w}.

\bibitem{44}
A.~Ghosh, H.~Kumar, P.~S. Sastry, Robust {L}oss {F}unctions under {L}abel {N}oise for {D}eep {N}eural {N}etworks, in: Proceedings of the Thirty-First {AAAI} Conference on Artificial Intelligence, {AAAI}, San Francisco, California, {USA}, February 4-9, 2017, pp. 1919--1925, \href{https://doi.org/10.1609/aaai.v31i1.10894}{https://doi.org/10.1609/aaai.v31i1.10894}.

\bibitem{45}
E.~Bauer, R.~Kohavi, An {E}mpirical {C}omparison of {V}oting {C}lassification {A}lgorithms: {B}agging, {B}oosting, and {V}ariants, Mach. Learn. 36~(1) (1999) 105--139, \href{https://doi.org/10.1023/A:1007515423169}{https://doi.org/10.1023/A:1007515423169}.

\bibitem{41}
C.~Willmott, K.~Matsuura, Advantages of the mean absolute error ({MAE}) over the root mean square error ({RMSE}) in assessing average model performance, Climate Research. 30~(1) (2005) 79--82, \href{https://doi.org/10.3354/cr030079}{https://doi.org/10.3354/cr030079}.

\bibitem{46}
I.~Avcibas, B.~Sankur, K.~Sayood, Statistical evaluation of image quality measures, J. Electronic Imaging. 11~(2) (2002) 206--223, \href{https://doi.org/10.1117/1.1455011}{https://doi.org/10.1117/1.1455011}.

\bibitem{40}
H.~R.~S. Zhou~Wang, A. C.~Bovik, E.~P. Simoncelli, Image quality assessment: from error visibility to structural similarity, {IEEE} Trans. Image Process. 13~(4) (2004) 600--612, \href{https://doi.org/10.1109/TIP.2003.819861}{https://doi.org/10.1109/TIP.2003.819861}.

\bibitem{43}
M.~Heusel, H.~Ramsauer, T.~Unterthiner, B.~Nessler, S.~Hochreiter, {GAN}s {T}rained by a {T}wo {T}ime-{S}cale {U}pdate {R}ule {C}onverge to a {L}ocal {N}ash {E}quilibrium, in: Proceedings of Advances in Neural Information Processing Systems 30: Annual Conference on Neural Information Processing Systems, {NIPS}, Long Beach, CA, {USA}, December 4-9, 2017, pp. 6626--6637.

\bibitem{42}
R.~Zhang, P.~Isola, A.~A. Efros, E.~Shechtamn, O.~Wang, The {U}nreasonable {E}ffectiveness of {D}eep {F}eatures as a {P}erceptual {M}etric, in: Proceedings of {IEEE} Conference on Computer Vision and Pattern Recognition, {CVPR}, Salt Lake City, UT, USA, June 18-22, 2018, pp. 586--595, \href{https://doi.org/10.1109/CVPR.2018.00068}{https://doi.org/10.1109/CVPR.2018.00068}.

\bibitem{130}
C.~Wang, M.~Zhou, T.~Ge, Y.~Jiang, H.~Bao, W.~Xu, C{F}-{F}ont: {C}ontent {F}usion for {F}ew-shot {F}ont {G}eneration, in: Proceedings of {IEEE/CVF} Conference on Computer Vision and Pattern Recognition, {CVPR}, Vancouver, BC, Canada, June 17-24, 2023, pp. 1858--1867, \href{https://doi.org/10.1109/CVPR52729.2023.00185}{https://doi.org/10.1109/CVPR52729.2023.00185}.

\bibitem{146}
Y.~Kong, C.~Luo, W.~Ma, et~al., {L}ook {C}loser to {S}upervise {B}etter: {O}ne-{S}hot {F}ont {G}eneration via {C}omponent-{B}ased discriminator, in: Proceedings of {IEEE/CVF} Conference on Computer Vision and Pattern Recognition, {CVPR}, New Orleans, LA, USA, June 18-24, 2022, pp. 13472--13481, \href{https://doi.org/10.1109/CVPR52688.2022.01312}{https://doi.org/10.1109/CVPR52688.2022.01312}.

\bibitem{47}
B.~Fu, F.~Yu, A.~Liu, et~al., Generate {L}ike {E}xperts: {M}ulti-{S}tage {F}ont {G}eneration by {I}ncorporating {F}ont {T}ransfer {P}rocess into {D}iffusion {M}odels, in: Proceedings of the IEEE/CVF Conference on Computer Vision and Pattern Recognition, {CVPR}, Seattle WA, USA, June 17-21, 2024, pp. 6892--6901.

\bibitem{48}
\href{https://github.com/kaonashi-tyc/Rewrite}{https://github.com/kaonashi-tyc/Rewrite}.

\bibitem{51}
P.~Isola, J.~Zhu, et~al., Image-to-image translation with conditional adversarial networks, in: Proceedings of {IEEE} Conference on Computer Vision and Pattern Recognition, {CVPR}, Honolulu, HI, USA, July 21-26, 2017, pp. 5967--5976, \href{https://doi.org/10.1109/CVPR.2017.632}{https://doi.org/10.1109/CVPR.2017.632}.

\bibitem{50}
\href{https://github.com/kaonashi-tyc/zi2zi}{https://github.com/kaonashi-tyc/zi2zi}.

\bibitem{55}
Y.~Jiang, Z.~Lian, Y.~Tang, J.~Xiao, {DCF}ont: an end-to-end deep {C}hinese font generation system, in: Proceedings of {SIGGRAPH} Asia Technical Briefs, Bangkok, Thailand, November 27-30, 2017, pp. 1--4, \href{https://doi.org/10.1145/3145749.3149440}{https://doi.org/10.1145/3145749.3149440}.

\bibitem{52}
J.~Chang, Y.~Gu, Y.~Zhang, Y.~Wang, Chinese {H}andwriting {I}mitation with {H}ierarchical {G}enerative {A}dversarial {N}etwork, in: Proceedings of British Machine Vision Conference, {BMVC}, Newcastle, UK, September 3-6, 2018, pp. 1--12.

\bibitem{56}
P.~Lyu, X.~Bai, C.~Yao, Z.~Zhu, T.~Huang, W.~Liu, Auto-{E}ncoder {G}uided {GAN} for {C}hinese {C}alligraphy {S}ynthesis, in: Proceedings of the 14th {IAPR} International Conference on Document Analysis and Recognition, {ICDAR}, Kyoto, Japan, November 9-15, 2017, pp. 1095--1100, \href{https://doi.org/10.1109/ICDAR.2017.181}{https://doi.org/10.1109/ICDAR.2017.181}.

\bibitem{59}
D.~Sun, Q.~Zhang, J.~Yang, Pyramid {E}mbedded {G}enerative {A}dversarial {N}etwork for {A}utomated {F}ont {G}eneration, in: Proceedings of the 24th International Conference on Pattern Recognition, {ICPR}, Beijing, China, August 20-24, 2018, pp. 976--981, \href{https://doi.org/10.1109/ICPR.2018.8545701}{https://doi.org/10.1109/ICPR.2018.8545701}.

\bibitem{60}
O.~Ronneberger, P.~Fischer, T.~Brox, U-{N}et: {C}onvolutional {N}etworks for {B}iomedical {I}mage {S}egmentation, in: Proceedings of Medical Image Computing and Computer-Assisted Intervention - {MICCAI} 18th International Conference Munich, Germany, October 5-9, 2015, pp. 234--241, \href{https://doi.org/10.1007/978-3-319-24574-4\_28}{https://doi.org/10.1007/978-3-319-24574-4\_28}.

\bibitem{69}
Z.~Zhang, X.~Zhou, M.~Qin, X.~Chen, Chinese character style transfer based on multi-scale {GAN}, Signal Image Video Process. 16~(2) (2022) 559–567, \href{https://doi.org/10.1007/s11760-021-02000-6}{https://doi.org/10.1007/s11760-021-02000-6}.

\bibitem{61}
Q.~Zhang, Font style transfer algorithms based on generative adversarial nets, Dalian Maritime University, Liaoning, 2019.

\bibitem{80}
P.~Wang, P.~Chen, Y.~Yuan, D.~Liu, Z.~Huang, X.~Hou, et~al., Understanding {C}onvolution for {S}emantic {S}egmentation, in: Proceedings of {IEEE} Winter Conference on Applications of Computer Vision, {WACV}, Lake Tahoe, NV, USA, March 12-15, 2018, pp. 1451--1460, \href{https://doi.org/10.1109/WACV.2018.00163}{https://doi.org/10.1109/WACV.2018.00163}.

\bibitem{62}
C.~Ren, S.~Lyu, H.~Zhan, Y.~Lu, {SAF}ont: {A}utomatic {F}ont {S}ynthesis using {S}elf-{A}ttention {M}echanisms, Aust. J. Intell. Inf. Process. Syst. 16~(2) (2019) 19–25.

\bibitem{72}
Y.~Miao, H.~Jia, K.~Tang, Chinese font migration combining local and global features learning, Pattern Anal Applic. 24 (2021) 1533–1547, \href{https://doi.org/10.1007/s10044-021-01003-w}{https://doi.org/10.1007/s10044-021-01003-w}.

\bibitem{58}
Y.~Guo, Z.~Lian, Y.~Tang, J.~Xiao, Creating {N}ew {C}hinese {F}onts based on {M}anifold {L}earning and {A}dversarial {N}etworks, in: Proceedings of 39th Annual Conference of the European Association for Computer Graphics, Eurographics - Short Papers, Delft, The Netherlands, April 16-20, 2018, pp. 61--64, \href{https://doi.org/10.2312/egs.20181045}{https://doi.org/10.2312/egs.20181045}.

\bibitem{53}
Z.~Lian, B.~Zhao, J.~Xiao, Automatic generation of large-scale handwriting fonts via style learning, in: Proceedings of SIGGRAPH ASIA Technical Briefs, Macao, December 5-8, 2016, pp. 1--4, \href{https://doi.org/10.1145/3005358.3005371}{https://doi.org/10.1145/3005358.3005371}.

\bibitem{63}
Y.~Jiang, Z.~Lian, Y.~Tang, J.~Xiao, {SCF}ont: {S}tructure-{G}uided {C}hinese {F}ont {G}eneration via {D}eep {S}tacked {N}etworks, in: Proceedings of the Thirty-Third {AAAI} Conference on Artificial Intelligence, {AAAI}, Honolulu, Hawaii, USA, January 27 - February 1, 2019, pp. 4015--4022, \href{https://doi.org/10.1609/aaai.v33i01.33014015}{https://doi.org/10.1609/aaai.v33i01.33014015}.

\bibitem{64}
Y.~Gao, Z.~Lian, Y.~Tang, J.~Xiao, Automatic {G}eneration of {C}hinese {V}ector {F}onts via {D}eep {L}ayout {I}nferring, in: Proceedings of {SIGGRAPH} Asia Technical Briefs, Brisbane, QLD, Australia, November 17-20, 2019, pp. 33--36, \href{https://doi.org/10.1145/3355088.3365142}{https://doi.org/10.1145/3355088.3365142}.

\bibitem{66}
S.~J. Wu, C.~Yang, J.~Y. Hsu, Calli{GAN}: {S}tyle and {S}tructure-aware {C}hinese {C}alligraphy {C}haracter {G}enerator, preprint at \href{https://arxiv.org/abs/2005.12500}{https://arxiv.org/abs/2005.12500} (2020).

\bibitem{67}
J.~Zhang, D.~Chen, G.~Han, G.~Li, J.~He, Z.~Liu, Z.~Ruan, {SSN}et: {S}tructure-{S}emantic {N}et for {C}hinese typography generation based on image translation, Neurocomputing. 371 (2020) 15--26, \href{https://doi.org/10.1016/j.neucom.2019.08.072}{https://doi.org/10.1016/j.neucom.2019.08.072}.

\bibitem{70}
C.~Wen, J.~Chang, Y.~Zhang, S.~Chen, Y.~Wang, M.~Han, Q.~Tian, Handwritten {C}hinese {F}ont {G}eneration with {C}ollaborative {S}troke {R}efinement, in: Proceedings of {IEEE} Winter Conference on Applications of Computer Vision, {WACV}, Waikoloa, HI, USA, January 3-8, 2021, pp. 3881--3890, \href{https://doi.org/10.1109/WACV48630.2021.00393}{https://doi.org/10.1109/WACV48630.2021.00393}.

\bibitem{71}
C.~Wang, Y.~Ding, Y.~Liu, G.~Zhan, Z.~Li, Chinese {F}ont {G}eneration from {S}troke {S}emantic and {A}ttention {M}echanism, Journal of Computer-Aided Design \& Computer Graphics. 34~(8) (2022) 1229--1237, \href{https://doi.org/10.3724/SP.J.1089.2022.19125}{https://doi.org/10.3724/SP.J.1089.2022.19125}.

\bibitem{75}
Z.~Lian, Y.~Gao, {CVF}ont: {S}ynthesizing {C}hinese {V}ector {F}onts via {D}eep {L}ayout {I}nferring, Comput. Graph. Forum. 41~(6) (2022) 212–225, \href{https://doi.org/10.1111/cgf.14580}{https://doi.org/10.1111/cgf.14580}.

\bibitem{77}
Y.~Liu, F.~binti Khalid, C.~Wang, et~al., An end-to-end {C}hinese font generation network with stroke semantics and deformable attention skip-connection, Expert Syst. Appl. 237 (2024) 1–13, \href{https://doi.org/10.1016/j.eswa.2023.121407}{https://doi.org/10.1016/j.eswa.2023.121407}.

\bibitem{88}
P.~Zhou, Z.~Zhao, K.~Zhang, C.~Li, C.~Wang, An end-to-end model for {C}hinese calligraphy generation, Multimed Tools Appl. 80~(5) (2021) 6737–6754, \href{https://doi.org/10.1007/s11042-020-09709-5}{https://doi.org/10.1007/s11042-020-09709-5}.

\bibitem{81}
B.~Chang, Q.~Zhang, S.~Pan, L.~Meng, Generating {H}andwritten {C}hinese {C}haracters {U}sing {C}ycle{GAN}, in: Proceedings of {IEEE} Winter Conference on Applications of Computer Vision, {WACV}, Lake Tahoe, NV, USA, March 12-15, 2018, pp. 199--207, \href{https://doi.org/10.1109/WACV.2018.00028}{https://doi.org/10.1109/WACV.2018.00028}.

\bibitem{109}
G.~Huang, Z.~Liu, et~al., {D}ensely {C}onnected {C}onvolutional {N}etworks, in: Proceedings of {IEEE} Conference on Computer Vision and Pattern Recognition, {CVPR}, Honolulu, HI, USA, July 21-26, 2017, pp. 2261--2269, \href{https://doi.org/10.1109/CVPR.2017.243}{https://doi.org/10.1109/CVPR.2017.243}.

\bibitem{82}
X.~Zhang, F.~Yin, Y.~Zhang, C.~Liu, Y.~Bengio, Drawing and {R}ecognizing {C}hinese {C}haracters with {R}ecurrent {N}eural {N}etwork, {IEEE} Trans. Pattern Anal. Mach. Intell. 40~(4) (2018) 849--862, \href{https://doi.org/10.1109/TPAMI.2017.2695539}{https://doi.org/10.1109/TPAMI.2017.2695539}.

\bibitem{83}
S.~Tang, Z.~Xia, Z.~Lian, Y.~Tang, J.~Xiao, Font{RNN}: {G}enerating {L}arge-scale {C}hinese {F}onts via {R}ecurrent {N}eural {N}etwork, Comput. Graph. Forum. 38~(7) (2019) 567--577, \href{https://doi.org/10.1111/cgf.13861}{https://doi.org/10.1111/cgf.13861}.

\bibitem{84}
X.~Liu, G.~Meng, S.~Xiang, C.~Pan, Font{GAN}: {A} {U}nified {G}enerative {F}ramework for {C}hinese {C}haracter {S}tylization and {D}e-stylization, preprint at \href{http://arxiv.org/abs/1910.12604}{http://arxiv.org/abs/1910.12604} (2019).

\bibitem{86}
Y.~Zhang, Generating Handwritten Chinese Characters with GANs, East China Normal University, Shanghai, 2019.

\bibitem{89}
L.~Wu, X.~Chen, et~al., Multitask {A}dversarial {L}earning for {C}hinese {F}ont {S}tyle {T}ransfer, in: Proceedings of International Joint Conference on Neural Networks, {IJCNN}, Glasgow, United Kingdom, July 19-24, 2020, pp. 1--8, \href{https://doi.org/10.1109/IJCNN48605.2020.9206851}{https://doi.org/10.1109/IJCNN48605.2020.9206851}.

\bibitem{90}
W.~Fan, Research on Chinese character generation technology based on generation adversarial network, Nanning Normal University, Guangxi, 2020.

\bibitem{91}
H.~Zhang, Research on Chinese Character Generation Method Based on Generative Adversarial Networks, Tianjin Normal University, Tianjin, 2020.

\bibitem{92}
Y.~Gao, J.~Wu, {GAN}-{B}ased {U}npaired {C}hinese {C}haracter {I}mage {T}ranslation via {S}keleton {T}ransformation and {S}troke {R}endering, in: Proceedings of The Thirty-Fourth {AAAI} Conference on Artificial Intelligence, {AAAI}, New York, NY, USA, February 7-12, 2020, pp. 646--653, \href{https://doi.org/10.1609/aaai.v34i01.5405}{https://doi.org/10.1609/aaai.v34i01.5405}.

\bibitem{94}
Y.~Lin, H.~Yuan, L.~Lin, {C}hinese {T}ypography {T}ransfer {M}odel {B}ased on {G}enerative {A}dversarial {N}etwork, in: Proceedings of 2020 Chinese Automation Congress (CAC), Shanghai, China, November 6-8, 2020, pp. 7005--7010, \href{https://doi.org/10.1109/CAC51589.2020.9326672}{https://doi.org/10.1109/CAC51589.2020.9326672}.

\bibitem{29}
Y.~Xiao, W.~Lei, L.~Lu, X.~Chang, X.~Zheng, X.~Chen, {CS-GAN:} {C}ross-{S}tructure {G}enerative {A}dversarial {N}etworks for {C}hinese calligraphy translation, Knowl. Based Syst. 229 (2021) 1--10, \href{https://doi.org/10.1016/j.knosys.2021.107334}{https://doi.org/10.1016/j.knosys.2021.107334}.

\bibitem{54}
C.~Mu, {R}esearch and {I}mplementation of {C}alligraphy {C}hinese {C}haracter {G}eneration {B}ased on {S}tyle {T}ransfer {T}echnology, University of Electronic Science and Technology of China, Sichuan, 2021.

\bibitem{95}
M.~Xue, J.~Du, J.~Zhang, Z.~Wang, B.~Wang, B.~Ren, Radical {C}omposition {N}etwork for {C}hinese {C}haracter {G}eneration, in: Proceedings of the 16th International Conference on Document Analysis and Recognition, {ICDAR}, Lausanne, Switzerland, September 5-10, 2021, pp. 252--267, \href{https://doi.org/10.1007/978-3-030-86549-8\_17}{https://doi.org/10.1007/978-3-030-86549-8\_17}.

\bibitem{98}
A.~U. Hassan, H.~Ahmed, J.~Choi, Unpaired font family synthesis using conditional generative adversarial networks, Knowl. Based Syst. 229 (2021) 1--12, \href{https://doi.org/10.1016/j.knosys.2021.107304}{https://doi.org/10.1016/j.knosys.2021.107304}.

\bibitem{96}
X.~Liu, G.~Meng, J.~Chang, R.~Hu, S.~Xiang, C.~Pan, Decoupled {R}epresentation {L}earning for {C}haracter {G}lyph {S}ynthesis, {IEEE} Trans. Multim. 24 (2022) 1787--1799, \href{https://doi.org/10.1109/TMM.2021.3072449}{https://doi.org/10.1109/TMM.2021.3072449}.

\bibitem{101}
J.~Zhou, Y.~Wang, Y.~Yuan, Q.~Huang, J.~Zeng, {SGCE}-{F}ont: {S}keleton {G}uided {C}hannel {E}xpansion for {C}hinese {F}ont {G}eneration, preprint at \href{http://arxiv.org/abs/2211.14475}{http://arxiv.org/abs/2211.14475} (2022).

\bibitem{102}
J.~Zeng, Q.~Chen, M.~Wang, Self-supervised {C}hinese font generation based on square-block transformation, Sci Sin Inform. 52~(1) (2022) 145--159, \href{https://doi.org/10.1360/SSI-2021-0056}{https://doi.org/10.1360/SSI-2021-0056}.

\bibitem{103}
Q.~Liao, G.~Xia, Z.~Wang, Calliffusion: {C}hinese {C}alligraphy {G}eneration and {S}tyle {T}ransfer with {D}iffusion {M}odeling, preprint at \href{http://arxiv.org/abs/2305.19124}{http://arxiv.org/abs/2305.19124} (2023).

\bibitem{105}
X.~Ye, H.~Zhang, L.~Yang, J.~Du, {R}adical {C}onstraint-{B}ased {G}enerative {A}dversarial {N}etwork for {H}andwritten {C}hinese {C}haracter {G}eneration, Comput. Informatics. 43~(2) (2024) 482--504, \href{https://doi.org/10.31577/cai\_2024\_2\_482}{https://doi.org/10.31577/cai\_2024\_2\_482}.

\bibitem{166}
J.~Chung, {\c{C}}.~G{\"{u}}l{\c{c}}ehre, K.~Cho, Y.~Bengio, {E}mpirical {E}valuation of {G}ated {R}ecurrent {N}eural {N}etworks on {S}equence {M}odeling, preprint at \href{https://arxiv.org/abs/1412.3555}{https://arxiv.org/abs/1412.3555} (2014).

\bibitem{110}
Y.~Zhang, Y.~Zhang, W.~Cai, {S}eparating {S}tyle and {C}ontent for {G}eneralized {S}tyle {T}ransfer, in: Proceedings of {IEEE} Conference on Computer Vision and Pattern Recognition, {CVPR}, Salt Lake City, UT, USA, June 18-22, 2018, pp. 8447--8455, \href{https://doi.org/10.1109/CVPR.2018.00881}{https://doi.org/10.1109/CVPR.2018.00881}.

\bibitem{111}
H.~Jiang, G.~Yang, K.~Huang, R.~Zhang, {W}-{N}et: {O}ne-{S}hot {A}rbitrary-{S}tyle {C}hinese {C}haracter {G}eneration with {D}eep {N}eural {N}etworks, in: Proceedings of Neural Information Processing - 25th International Conference, {ICONIP}, Siem Reap, Cambodia, December 13-16, 2018, pp. 483--493, \href{https://doi.org/10.1007/978-3-030-04221-9\_43}{https://doi.org/10.1007/978-3-030-04221-9\_43}.

\bibitem{112}
X.~Yan, Y.~Wang, R.~Yi, Z.~Sun, Y.~Liu, {S}tar{F}ont: {E}nabling {F}ont {C}ompletion {B}ased on few {S}hots {E}xamples, in: Proceedings of the 3rd International Conference on Advances in Artificial Intelligence, {ICAAI}, Istanbul, Turkey, October 26-28, 2019, pp. 1--8, \href{https://doi.org/10.1145/3369114.3369115}{https://doi.org/10.1145/3369114.3369115}.

\bibitem{114}
S.~Yang, J.~Liu, W.~Wang, Z.~Guo, {TET-GAN:} {T}ext {E}ffects {T}ransfer via {S}tylization and {D}estylization, in: Proceedings of the Thirty-Third {AAAI} Conference on Artificial Intelligence, {AAAI}, Honolulu, Hawaii, USA, January 27 - February 1, 2019, pp. 1238--1245, \href{https://doi.org/10.1609/aaai.v33i01.33011238}{https://doi.org/10.1609/aaai.v33i01.33011238}.

\bibitem{113}
B.~Zhao, J.~Tao, M.~Yang, Z.~Tian, C.~Fan, Y.~Bai, Deep imitator: {H}andwriting calligraphy imitation via deep attention networks, Pattern Recognit. 104 (2020) 1--14, \href{https://doi.org/10.1016/j.patcog.2019.107080}{https://doi.org/10.1016/j.patcog.2019.107080}.

\bibitem{120}
Z.~Lai, C.~Tang, J.~Lv, Arbitrary {C}hinese {F}ont {G}eneration from a {S}ingle {R}eference, in: Proceedings of International Joint Conference on Neural Networks, {IJCNN}, Glasgow, United Kingdom, July 19-24, 2020, pp. 1--7, \href{https://doi.org/10.1109/IJCNN48605.2020.9206919}{https://doi.org/10.1109/IJCNN48605.2020.9206919}.

\bibitem{121}
H.~Aoki, K.~Tsubota, H.~Ikuta, K.~Aizawa, Few-{S}hot {F}ont {G}eneration with {D}eep {M}etric {L}earning, in: Proceedings of 25th International Conference on Pattern Recognition, {ICPR}, Milan, Italy, January 10-15, 2020, pp. 8539--8546, \href{https://doi.org/10.1109/ICPR48806.2021.9412254}{https://doi.org/10.1109/ICPR48806.2021.9412254}.

\bibitem{122}
A.~Zhu, X.~Lu, X.~Bai, S.~Uchida, B.~K. Iwana, S.~Xiong, {F}ew-{S}hot {T}ext {S}tyle {T}ransfer via {D}eep {F}eature {S}imilarity, {IEEE} Trans. Image Process. 29 (2020) 6932--6946, \href{https://doi.org/10.1109/TIP.2020.2995062}{https://doi.org/10.1109/TIP.2020.2995062}.

\bibitem{123}
Y.~Wang, Y.~Gao, Z.~Lian, {A}ttribute2{F}ont: creating fonts you want from attributes, {ACM} Trans. Graph. 39~(4) (2020) 1--15, \href{https://doi.org/10.1145/3386569.3392456}{https://doi.org/10.1145/3386569.3392456}.

\bibitem{116}
X.~Chen, L.~Wu, M.~He, et~al., {MLF}ont: {F}ew-{S}hot {C}hinese {F}ont {G}eneration via {D}eep {M}eta-{L}earning, in: Proceedings of International Conference on Multimedia Retrieval, {ICMR}, Taipei, Taiwan, China, August 21-24, 2021, pp. 37--45, \href{https://doi.org/10.1145/3460426.3463606}{https://doi.org/10.1145/3460426.3463606}.

\bibitem{117}
C.~Finn, P.~Abbeel, S.~Levine, Model-{A}gnostic {M}eta-{L}earning for {F}ast {A}daptation of {D}eep {N}etworks, in: Proceedings of the 34th International Conference on Machine Learning, {ICML}, Sydney, NSW, Australia, August 6-11, 2017, pp. 1126--1135.

\bibitem{115}
X.~Chen, Few-shot Font Generation Based on Deep Learning, Shandong University, Shandong, 2022.

\bibitem{119}
Q.~Wen, S.~Li, et~al., Zi{GAN}: {F}ine-grained {C}hinese {C}alligraphy {F}ont {G}eneration via a {F}ew-shot {S}tyle {T}ransfer {A}pproach, in: Proceedings of {ACM} Multimedia Conference, Virtual Event, China, October 20 - 24, 2021, pp. 621--629, \href{https://doi.org/10.1145/3474085.3475225}{https://doi.org/10.1145/3474085.3475225}.

\bibitem{136}
K.~Baek, Y.~Choi, Y.~Uh, J.~Yoo, H.~Shim, Rethinking the {T}ruly {U}nsupervised {I}mage-to-{I}mage {T}ranslation, in: Proceedings of {IEEE/CVF} International Conference on Computer Vision, {ICCV}, Montreal, QC, Canada, October 10-17, 2021, pp. 14134--14143, \href{https://doi.org/10.1109/ICCV48922.2021.01389}{https://doi.org/10.1109/ICCV48922.2021.01389}.

\bibitem{124}
Y.~Xie, X.~Chen, L.~Sun, Y.~Lu, {DG}-{F}ont: {D}eformable {G}enerative {N}etworks for {U}nsupervised {F}ont {G}eneration, in: Proceedings of {IEEE} Conference on Computer Vision and Pattern Recognition, {CVPR}, virtual, June 19-25, 2021, pp. 5126--5136, \href{https://doi.org/10.1109/CVPR46437.2021.00509}{https://doi.org/10.1109/CVPR46437.2021.00509}.

\bibitem{167}
J.~Dai, H.~Qi, Y.~Xiong, {D}eformable {C}onvolutional {N}etworks, in: Proceedings of {IEEE} International Conference on Computer Vision, {ICCV} 2017, Venice, Italy, October 22-29, 2017, pp. 764--773, \href{https://doi.org/10.1109/ICCV.2017.89}{https://doi.org/10.1109/ICCV.2017.89}.

\bibitem{135}
X.~Chen, Y.~Xie, L.~Sun, Y.~Lu, {DGF}ont++: {R}obust {D}eformable {G}enerative {N}etworks for {U}nsupervised {F}ont {G}eneration, preprint at \href{https://arxiv.org/abs/2212.14742}{https://arxiv.org/abs/2212.14742} (2022).

\bibitem{126}
Y.~Zhang, J.~Man, P.~Sun, {MF}-{N}et: {A} {N}ovel {F}ew-shot {S}tylized {M}ultilingual {F}ont {G}eneration {M}ethod, in: Proceedings of the 30th {ACM} International Conference on Multimedia, Lisboa, Portugal, October 10 - 14, 2022, pp. 2088--2096, \href{https://doi.org/10.1145/3503161.3548414}{https://doi.org/10.1145/3503161.3548414}.

\bibitem{131}
A.~U. Hassan, I.~Memon, J.~Choi, Learning font-style space using style-guided discriminator for few-shot font generation, Expert Syst. Appl. 242 (2024) 1--11, \href{https://doi.org/10.1016/j.eswa.2023.122817}{https://doi.org/10.1016/j.eswa.2023.122817}.

\bibitem{128}
X.~He, M.~Zhu, N.~Wang, X.~Gao, H.~Yang, Few-shot {F}ont {G}eneration by {L}earning {S}tyle {D}ifference and {S}imilarity, preprint at \href{http://arxiv.org/abs/2301.10008}{http://arxiv.org/abs/2301.10008} (2023).

\bibitem{137}
B.~Mildenhall, P.~P. Srinivasan, M.~Tancik, J.~T. Barron, R.~Ramamoorthi, R.~Ng, Ne{RF}: {R}epresenting {S}cenes as {N}eural {R}adiance {F}ields for {V}iew {S}ynthesis, in: Proceedings of Computer Vision - {ECCV}- 16th European Conference, Glasgow, UK, August 23-28, 2020, pp. 405--421, \href{https://doi.org/10.1007/978-3-030-58452-8\_24}{https://doi.org/10.1007/978-3-030-58452-8\_24}.

\bibitem{129}
B.~Fu, J.~He, J.~Wang, Y.~Qiao, {N}eural {T}ransformation {F}ields for {A}rbitrary-{S}tyled {F}ont {G}eneration, in: Proceedings of {IEEE/CVF} Conference on Computer Vision and Pattern Recognition, {CVPR}, Vancouver, BC, Canada, June 17-24, 2023, pp. 22438--22447, \href{https://doi.org/10.1109/CVPR52729.2023.02149}{https://doi.org/10.1109/CVPR52729.2023.02149}.

\bibitem{134}
Q.~Jin, F.~He, W.~Tang, C{LF}-{N}et: {A} {F}ew-shot {C}ross-{L}anguage {F}ont {G}eneration {M}ethod, in: Proceedings of MultiMedia Modeling - 30th International Conference, {MMM}, Amsterdam, The Netherlands, January 29 - February 2, 2024, pp. 127--140, \href{https://doi.org/10.1007/978-3-031-53308-2\_10}{https://doi.org/10.1007/978-3-031-53308-2\_10}.

\bibitem{127}
B.~Fu, F.~Yu, A.~Liu, et~al., Generate {L}ike {E}xperts: {M}ulti-{S}tage {F}ont {G}eneration by {I}ncorporating {F}ont {T}ransfer {P}rocess into {D}iffusion {M}odels, in: Proceedings of {IEEE/CVF} Conference on Computer Vision and Pattern Recognition, {CVPR}, Seattle, WA, USA, June 16-22, 2024, pp. 6892--6901, \href{https://doi.org/10.1109/CVPR52733.2024.00658}{https://doi.org/10.1109/CVPR52733.2024.00658}.

\bibitem{172}
Y.~Liu, Y.~Ding, X.~Li, et~al., {U}nsupervised {F}ont {G}eneration {N}etwork {I}ntegrating {C}ontent and {S}tyle {R}epresentation, Journal of Computer-Aided Design \& Computer Graphics 37~(5) (2025) 865--876.

\bibitem{139}
Z.~Lian, B.~Zhao, X.~Chen, J.~Xiao, Easy{F}ont: {A} {S}tyle {L}earning-{B}ased {S}ystem to {E}asily {B}uild {Y}our {L}arge-{S}cale {H}andwriting {F}onts, {ACM} Trans. Graph. 38~(1) (2018) 1--18, \href{https://doi.org/10.1145/3213767}{https://doi.org/10.1145/3213767}.

\bibitem{140}
Y.~Huang, M.~He, L.~Jin, Y.~Wang, {RD-GAN:} {F}ew/{Z}ero-{S}hot {C}hinese {C}haracter {S}tyle {T}ransfer via {R}adical {D}ecomposition and {R}endering, in: Proceedings of Computer Vision - {ECCV}- 16th European Conference, Glasgow, UK, August 23-28, 2020, pp. 156--172, \href{https://doi.org/10.1007/978-3-030-58539-6\_10}{https://doi.org/10.1007/978-3-030-58539-6\_10}.

\bibitem{141}
S.~Tang, Z.~Lian, Write {L}ike {Y}ou: {S}ynthesizing {Y}our {C}ursive {O}nline {C}hinese {H}andwriting via {M}etric-based {M}eta {L}earning, Comput. Graph. Forum. 40~(2) (2021) 141--151, \href{https://doi.org/10.1111/cgf.142621}{https://doi.org/10.1111/cgf.142621}.

\bibitem{145}
W.~Liu, F.~Liu, F.~Ding, Q.~He, Z.~Yi, {XMP}-{F}ont: {S}elf-{S}upervised {C}ross-{M}odality {P}re-training for {F}ew-{S}hot {F}ont {G}eneration, in: Proceedings of {IEEE/CVF} Conference on Computer Vision and Pattern Recognition, {CVPR}, New Orleans, LA, USA, June 18-24, 2022, pp. 7895--7904, \href{https://doi.org/10.1109/CVPR52688.2022.00775}{https://doi.org/10.1109/CVPR52688.2022.00775}.

\bibitem{142}
Y.~Liu, Z.~Lian, Font{RL}: {C}hinese {F}ont {S}ynthesis via {D}eep {R}einforcement {L}earning, in: Proceedings of Thirty-Fifth {AAAI} Conference on Artificial Intelligence, {AAAI}, Virtual Event, February 2-9, 2021, pp. 2198--2206, \href{https://doi.org/10.1609/aaai.v35i3.16318}{https://doi.org/10.1609/aaai.v35i3.16318}.

\bibitem{143}
S.~Park, S.~Chun, J.~Cha, B.~Lee, H.~Shim, Few-shot {F}ont {G}eneration with {L}ocalized {S}tyle {R}epresentations and {F}actorization, in: Proceedings of Thirty-Fifth {AAAI} Conference on Artificial Intelligence, {AAAI}, Virtual Event, February 2-9, 2021, pp. 2393--2402, \href{https://doi.org/10.1609/aaai.v35i3.16340}{https://doi.org/10.1609/aaai.v35i3.16340}.

\bibitem{144}
S.~Park, S.~Chun, J.~Cha, B.~Lee, H.~Shim, Multiple {H}eads are {B}etter than {O}ne: {F}ew-shot {F}ont {G}eneration with {M}ultiple {L}ocalized {E}xperts, in: Proceedings of {IEEE/CVF} International Conference on Computer Vision, {ICCV}, Montreal, QC, Canada, October 10-17, 2021, pp. 13880--13889, \href{https://doi.org/10.1109/ICCV48922.2021.01364}{https://doi.org/10.1109/ICCV48922.2021.01364}.

\bibitem{173}
W.~Wang, D.~Sun, J.~Zhang, L.~Gao, {MX}-{F}ont++: {M}ixture of {H}eterogeneous {A}ggregation {E}xperts for {F}ew-shot {F}ont {G}eneration, in: Proceedings of IEEE International Conference on Acoustics, Speech and Signal Processing (ICASSP), Hyderabad, India, April 6-11, 2025, pp. 1--5, \href{https://doi.org/10.1109/ICASSP49660.2025.10888465}{https://doi.org/10.1109/ICASSP49660.2025.10888465}.

\bibitem{148}
L.~Tang, Y.~Cai, J.~Liu, Z.~Hong, M.~Gong, et~al., Few-{S}hot {F}ont {G}eneration by {L}earning {F}ine-{G}rained {L}ocal {S}tyles, in: Proceedings of {IEEE/CVF} Conference on Computer Vision and Pattern Recognition, {CVPR}, New Orleans, LA, USA, June 18-24, 2022, pp. 7885--7894, \href{https://doi.org/10.1109/CVPR52688.2022.00774}{https://doi.org/10.1109/CVPR52688.2022.00774}.

\bibitem{153}
M.~Zhao, X.~Qi, Z.~Hu, et~al., Calligraphy {F}ont {G}eneration via {E}xplicitly {M}odeling {L}ocation-{A}ware {G}lyph {C}omponent {D}eformations, {IEEE} Trans. Multim. 26 (2024) 5939--5950, \href{https://doi.org/10.1109/TMM.2023.3342690}{https://doi.org/10.1109/TMM.2023.3342690}.

\bibitem{149}
S.~Yuan, R.~Liu, M.~Chen, B.~Chen, et~al., {SE-GAN:} {S}keleton {E}nhanced {G}an-{B}ased {M}odel for {B}rush {H}andwriting {F}ont {G}eneration, in: Proceedings of {IEEE} International Conference on Multimedia and Expo, {ICME}, Taipei, Taiwan, China, July 18-22, 2022, pp. 1--6, \href{https://doi.org/10.1109/ICME52920.2022.9859964}{https://doi.org/10.1109/ICME52920.2022.9859964}.

\bibitem{150}
Y.~Liu, Z.~Lian, Font{T}ransformer: {F}ew-shot {H}igh-resolution {C}hinese {G}lyph {I}mage {S}ynthesis via {S}tacked {T}ransformers, Pattern Recognit. 141 (2023) 1--16, \href{https://doi.org/10.1016/j.patcog.2023.109593}{https://doi.org/10.1016/j.patcog.2023.109593}.

\bibitem{154}
Y.~Su, X.~Chen, L.~Wu, X.~Meng, Learning {C}omponent-{L}evel and {I}nter-{C}lass {G}lyph {R}epresentation for {F}ew-shot {F}ont {G}eneration, in: Proceedings of {IEEE} International Conference on Multimedia and Expo, {ICME}, Brisbane, Australia, July 10-14, 2023, pp. 738--743, \href{https://doi.org/10.1109/ICME55011.2023.00132}{https://doi.org/10.1109/ICME55011.2023.00132}.

\bibitem{155}
Y.~Zhang, Y.~Song, A.~Li, {H}andwritten {C}hinese {C}haracter {G}eneration via {E}mbedding, {D}ecomposition and {D}iscrimination, in: Proceedings of International Joint Conference on Neural Networks, {IJCNN}, Gold Coast, Australia, June 18-23, 2023, pp. 1--7, \href{https://doi.org/10.1109/IJCNN54540.2023.10191803}{https://doi.org/10.1109/IJCNN54540.2023.10191803}.

\bibitem{156}
G.~Dai, Y.~Zhang, Q.~Wang, et~al., Disentangling {W}riter and {C}haracter {S}tyles for {H}andwriting {G}eneration, in: Proceedings of {IEEE/CVF} Conference on Computer Vision and Pattern Recognition, {CVPR}, Vancouver, BC, Canada, June 17-24, 2023, pp. 5977--5986, \href{https://doi.org/10.1109/CVPR52729.2023.00579}{https://doi.org/10.1109/CVPR52729.2023.00579}.

\bibitem{159}
Y.~Liu, Z.~Lian, Deep{C}alli{F}ont: {F}ew-shot {C}hinese {C}alligraphy {F}ont {S}ynthesis by {I}ntegrating {D}ual-{M}odality {G}enerative {M}odels, in: Proceedings of Thirty-Eighth {AAAI} Conference on Artificial Intelligence, {AAAI}, Vancouver, Canada, February 20-27, 2024, pp. 3774--3782, \href{https://doi.org/10.1609/aaai.v38i4.28168}{https://doi.org/10.1609/aaai.v38i4.28168}.

\bibitem{162}
J.~Xiong, Y.~Wang, J.~Zeng, C{LIP-F}ont: {S}ementic {S}elf-{S}upervised {F}ew-{S}hot {F}ont {G}eneration with {CLIP}, in: Proceedings of {IEEE} International Conference on Acoustics, Speech and Signal Processing, {ICASSP}, Seoul, Republic of Korea, April 14-19, 2024, pp. 3620--3624, \href{https://doi.org/10.1109/ICASSP48485.2024.10447490}{https://doi.org/10.1109/ICASSP48485.2024.10447490}.

\bibitem{164}
L.~Zhang, Y.~Zhu, A.~Benarab, et~al., {DP-F}ont: {C}hinese {C}alligraphy {F}ont {G}eneration {U}sing {D}iffusion {M}odel and {P}hysical {I}nformation {N}eural {N}etwork, in: Proceedings of the Thirty-Third International Joint Conference on Artificial Intelligence, {IJCAI}, Jeju Island, South Korea, August 3-9, 2024, pp. 7796--7804, \href{https://doi.org/10.24963/ijcai.2024/863}{https://doi.org/10.24963/ijcai.2024/863}.

\bibitem{174}
Y.~Wang, K.~Xiong, Y.~Yuan, J.~Zeng, {E}dge{F}ont: {E}nhancing style and content representations in few-shot font generation with multi-scale edge self-supervision, Expert Syst. Appl. 262 (2025) 1--11, \href{https://doi.org/10.1016/J.ESWA.2024.125547}{https://doi.org/10.1016/J.ESWA.2024.125547}.

\bibitem{74}
W.~Chen, J.~Su, W.~Song, et~al., Quality evaluation methods of handwritten {C}hinese characters: a comprehensive survey, Multim. Syst. 30~(4) (2024) 1--29, \href{https://doi.org/10.1007/s00530-024-01396-8}{https://doi.org/10.1007/s00530-024-01396-8}.

\end{thebibliography}

\end{document}